\pdfoutput=1
\documentclass{article}

    \PassOptionsToPackage{numbers, compress}{natbib}
     \usepackage[preprint]{neurips_2023}

\usepackage[utf8]{inputenc} 
\usepackage[T1]{fontenc}

\usepackage{csquotes}
\usepackage{hyperref}       
\usepackage{url}            
\usepackage{booktabs}       
\usepackage{nicefrac}       
\usepackage{xcolor}         
\usepackage{multirow}
\usepackage{graphicx} 
\usepackage{amsmath} 
\usepackage{subcaption}
\usepackage{tabularx}
\usepackage{enumitem}
\usepackage{makecell}
\usepackage{enumitem}
\usepackage{mathrsfs}
\usepackage{dutchcal}
\usepackage {setspace}
\usepackage{amsmath, amssymb}

\usepackage{algorithm}
\usepackage{algpseudocode}

\def\rti{\textsf{RTI}{}}

\def\data{$({\mathbf{prompt}},{\mathbf{p}},{q},{o},{a})$}

\title{Assessing Hidden Risks of LLMs: An Empirical Study on Robustness, Consistency, and Credibility}

\author{%
   \makecell{Wentao Ye\textsuperscript{1}, Mingfeng Ou\textsuperscript{1}, Tianyi Li\textsuperscript{1}, Yipeng Chen\textsuperscript{1}, Xuetao Ma\textsuperscript{2}, Yifan Yanggong\textsuperscript{2}, \\Sai Wu\textsuperscript{1}, Jie Fu\textsuperscript{3}, Gang Chen\textsuperscript{1}, Haobo Wang\textsuperscript{1}, Junbo Zhao\textsuperscript{1}}\\
   \textsuperscript{1}Zhejiang University\\ 
   \textsuperscript{2}ZhongHao XinYing (Hangzhou) Technology Co., Ltd.\\
   \textsuperscript{3}Beijing Academy of Artificial Intelligence 
   }

\begin{document}

\maketitle

\begin{abstract}
The recent popularity of large language models (LLMs) has brought a significant impact to boundless fields, particularly through their open-ended ecosystem such as the APIs, open-sourced models, and plugins.
However, with their widespread deployment, there is a general lack of research that thoroughly discusses and analyzes the potential risks concealed. 
In that case, we intend to conduct a preliminary but pioneering study covering the robustness, consistency, and credibility of LLMs systems.
With most of the related literature in the era of LLM uncharted, we propose an automated workflow that copes with an upscaled number of queries/responses.
Overall, we conduct over a million queries to the mainstream LLMs including ChatGPT, LLaMA, and OPT.
Core to our workflow consists of a data primitive, followed by an automated interpreter that evaluates these LLMs under different adversarial metrical systems.
As a result, we draw several, and perhaps unfortunate, conclusions that are quite uncommon from this trendy community.
Briefly, they are: (i)-the minor but inevitable error occurrence in the user-generated query input may, by chance, cause the LLM to respond unexpectedly;
(ii)-LLMs possess poor consistency when processing semantically similar query input.
In addition, as a side finding, we find that ChatGPT is still capable to yield the correct answer even when the input is polluted at an extreme level.
While this phenomenon demonstrates the powerful memorization of the LLMs, it raises serious concerns about using such data for LLM-involved evaluation in academic development.
To deal with it, we propose a novel \emph{index} associated with a dataset that roughly decides the feasibility of using such data for LLM-involved evaluation.
Extensive empirical studies are tagged to support the aforementioned claims.

\end{abstract}

\section{Introduction}
\label{intro}

In recent few months, the LLMs --- in particular ChatGPT --- have swept the world by causing huge influence on numerous domains extending from the computer science community.
Assisted by the WebUI \footnote {https://chat.openai.com/}, open-source models \citet{touvron2023llama}, APIs \footnote {https://platform.openai.com/docs/api-reference} or the ecosystem Plugins\footnote{https://openai.com/blog/chatgpt-plugins}, these LLMs successfully immerse into everyone's life, worldwide.

This success of LLMs is truly unprecedented, and even rare in the entire spectrum of development of the technology.
This gives rise to various evaluation efforts surrounding large models\citep{zhong2023can,qin2023chatgpt,huang2023chatgpt,kocon2023chatgpt,yang2023exploring}, which encounter numerous challenges: (i)-The complexity of LLM's outputs necessitates heavy reliance on human evaluation, hindering large-scale assessments; (ii)-Unlike traditional NLP models, current LLMs are widely deployed and exposed, posing additional potential risks hard to be detected through existed NLP capability evaluation; (iii)-The massive and unknown training data of LLMs present a pressing issue in selecting trustworthy evaluation data.

Driven by this, we fully consider the issues that are commonly encountered in relevant scenarios but have been rarely addressed in previous works. And we propose an automated workflow to conduct a systematic study of the LLMs covering three new terms of \textbf{robustness}, \textbf{consistency}, and \textbf{credibility}, as a pioneer attempt.    
\begin{itemize}
    \item the \textbf{robustness} of the LLM refers to the malicious queries that are made intentionally or unintentionally. Indeed, one can cast this problem as a conventional threat model of adversarial examples for NLP. Rather, in this work we delimit the robustness of LLMs against the conventional counterpart, in particular matching the threat model to the realistic deployment of the LLMs.
    \item the \textbf{consistency} of the APIs. This is a new concept we intend to promote. Briefly, we try to quantitatively measure the distinction of the LLMs processing two semantically homogenous query inputs. To do that, we propose a novel threat model coupled with two LLM-adapted attacking means.
    \item the \textbf{credibility} of the LLM evaluation. This measurement is perhaps most related to the academic community centered around the LLMs. This line is very much alerted by the following phenomenon: on certain datasets, ChatGPT is still able to spit out the correct answer, \textbf{even when we pollute the passage information completely}! On one hand, this ``blind-eye'' QA capacity of ChatGPT can be deemed as the LLM manages to memorize the provided datasets. Despite that, we urge the community to take caution when using these datasets to evaluate any potential compositional framework that composes an LLM within.
\end{itemize}

\begin{figure}[tb]
  \centering
    \includegraphics[width=\columnwidth]{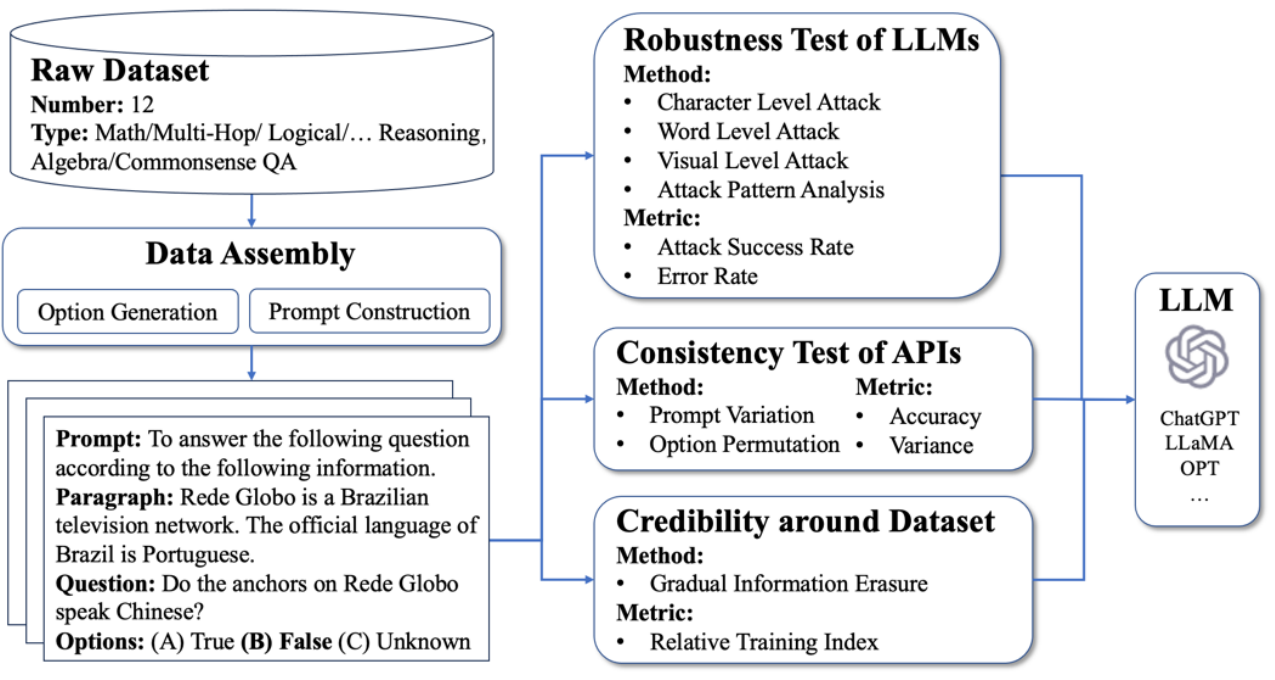}
  \caption{Overview of Evaluation Framework.}
  \label{fig:framework}
\end{figure}

In spite of the concise concepts we list above, there are several must-do steps to reach the evaluation of LLMs. Hereby, we briefly introduce our overall workflow.
\paragraph{Overall workflow}
First, we utilize the \texttt{gpt-3.5-turbo} API (the ChatGPT-level API), and the open-sourced LLaMA and OPT models as the main backbones for this study.
Notably, to attain the statistically sound results, we conduct over \textbf{one million} queries (around 0.7 billion tokens\footnote{https://openai.com/pricing} on ChatGPT API) for each of the LLMs.

Then, to cope with the scaled query-response pairs, an automated interpreter is required.
To do that, we devise a universal \emph{data primitive} that composes of five parts: \data, where five symbols in order indicate prompt, passage, question, confusion options, and the corresponding answer.
Upon 12 publicly available datasets, we manage to convert each data point to this form of primitive.
Noted, the nature of the primitive is a QA form with the answers being a multi-option selection problem. By regularizing the output from the LLM, we manage to process the scaled responses parallelly.

At last, centered around the three aspects of LLM evaluation, we respectively propose two threat models and five attack schemes with an index number associated with the dataset itself.
Without much detail, the two threat models point to robustness and consistency respectively.
For the five attack schemes, unlike the conventional NLP adversarial attacks \citet{jia2017adversarial}, we purposely structurize them such that they correspond to several specific usages of the LLM API/models.
The index --- dubbed as \rti{} --- is associated with the dataset. \rti{} provides a reference index to quantitatively measure the probability that the dataset has been memorized by the model. In that way, \rti{} can be viewed as a measure of the dataset's credibility, helping us decide whether to use this dataset in the evaluation.

\paragraph{Take-away lessons}.
With empirical study and analysis, our concerns can be summarized as follows.
\begin{enumerate}
    \item on the \textbf{robustness} of the LLMs, attacks ---especially those that are closely correlated to the usage of LLMs --- may pose a hazardous potential, such as the visual attacks for OCR-ed API input or typo error for ASR-ed input. 
    Moreover, the developers using ChatGPT ought to be careful with the input contents, due to that some minor but structured changes may drastically shift the model/API output unexpectedly, e.g. by no less than \textbf{27\%} as per our test for ChatGPT.
    \item from the \textbf{consistency} evaluation of the LLMs, we take ChatGPT as an example. Unfortunately, we find that with semantically identical query input, ChatGPT's accuracy of the responses at average fluctuates by 3.2\% in our testing protocol. Further noted, the change to the input is mostly in the aspects of grammatical expression, writing habits and etc., which happens quite commonly considering the vast deployment of these LLMs.
    \item \rti{}: \rti{} measures the relative probability that the dataset has been memorized. In other words, \rti{} is an index that characterizes if a provided dataset is suitable or credible for the LLM-involved evaluation process. We hope this metric can benefit the academic community towards developing a better suitable evaluation ecosystem in the era of LLMs.
\end{enumerate}

We open-source our datasets and samples at \url{https://github.com/yyy01/LLMRiskEval_RCC} for further development.

\section{Related Work}

\subsection{Adversarial Attack in NLP}

In traditional NLP tasks, there have been various works studying the adversarial attack models.
Due to the non-differential nature of the text data, the attacks are mostly conducted in a block-box threat model.
These attacks widely cover character-level manipulation~\citep{gao2018black, belinkov2017synthetic, heigold2017robust, eger2019text}, word-level manipulation~\citep{papernot2016crafting, alzantot2018generating, zhao2017generating}, and sentence-level manipulation~\citep{iyyer2018adversarial, li2020generating}. Moreover, other works~\citep{li2018textbugger, zang2019word, ebrahimi2017hotflip, li2020bert, wallace2019universal, garg2020bae} are following a white-box fashion, requiring access to the target model's gradients, structure, or parameters. However, as the present ChatGPT version offers only interfaces, we can only employ black-box attacks for the evaluation.

\subsection{Adversarial Attacks on LLMs}
Previously, ~\citet{altinisik2022impact,moradi2021evaluating,stolfo2022causal,zhang2022interpreting} presented a variety of robustness assessing schemes for LLMs.
This work generally resorts to using manageable perturbation on the input to the LLMs, such as typos, entities swap, negations, sentence insertion, etc.

Despite that, most --- if not all --- of these approaches are limited to smaller LLMs such as BERT~\cite{devlin2018bert}, and XLNET~\cite{yang2019xlnet}.
Given the current terrain of rapidly developing and enlarging the LLMs, we argue that LLMs at the scale of ChatGPT need to be studied.
Rather, a recent work~\citep{wang2023robustness} presents that ChatGPT demonstrates a consistent advantage in most adversarial classification and translation tasks.

\subsection{Threat Model}
The concept of the threat model originates from computer security. It refers to the process of identifying, evaluating, and managing threats that a system may suffer from.
In NLP, threat models are widely used to help administrators identify possible threats, so as to take corresponding preventive actions.
\citet{brendel2018decisionbased}, \citet{neekhara2019adversarial}, \citet{hambardzumyan2021warp}, \citet{wallace2021universal} and \citet{madry2019deep} have conducted various work on threat models, towards black-box systems or APIs. However, most of these methods more focus on smaller Machine Learning models, whose impact on LLMs is limited.
With the popularity of ChatGPT, new threat models, more closely integrated with LLMs' application scenarios, are urgently required.

\subsection{LLMs' Evaluation: Reliable?}
Indeed, with the rise of ChatGPT and GPT-4, there have been various works coming to light in evaluating these LLMs on off-the-shelf datasets.
To name a few, \citet{zhong2023can,qin2023chatgpt,huang2023chatgpt,kocon2023chatgpt,yang2023exploring} focus on developing additional model.

However, as we mentioned, the evaluation protocol when involving a powerful LLM requires further scrutiny.
That is, our RTI index indicates a rough and relative rate that the fed testing samples were already memorized by the LLM.
Consequently, the isolated evaluation on the external modules besides the LLM might be skeptical because we show that a ChatGPT is capable to answer questions without the help of any other information --- for some datasets.

Indeed, our RTI is only capable to imply a relative score for the datasets, not the absolute one, without any further information about the dataset which the LLM was trained upon.
We refer the reader to Section~\ref{sec:metric} for detailed information.

\section{Method, Data, and Desiderata}
\label{Dataset and Method}

In this work, we elaborate on the details of our workflow.
In that, we begin with the data form construction that is catered to this particular task.
In what follows, we detail the specific gauging mechanisms for evaluating the robustness, consistency, and credibility of the current LLMs.
To wrap up, we formally provide the metric system for this study.

\subsection{Raw Dataset Curation}
We adopt 12 publicly available datasets for benchmarking.
Noted, these datasets mainly contain tasks of 
mathematical deduction, logical reasoning, commonsense understanding and etc.
The full setup is listed in Table~\ref{Dataset}.

Formally, for each individual data point among these datasets, we use a triplet form to represent it, as $(\mathbf{p},q,a)$ where the three symbols in order indicate passage, question, and the corresponding answer respectively.
Notably, data points in certain datasets --- e.g. AQuA, Creak in Table~\ref{Dataset} --- do not involve a preset passage.
For uniformity, we may still refer them to the triplet form while the $\mathbf{p}$ variable is proactively set to Null.

\begin{table}[htb]
  \caption{Datasets statistics. \emph{Answer Type} row refers to the types of answers given in datasets, where T/F means True or False, Number means Numerical value, Word means English words, Text means multi-words or sentences, and multi means multi-types.)}
  \label{Dataset}
  \centering
  \begin{small}
  \begin{tabular}{ll|ccc}
    \toprule
    \multirow{2}*{\textbf{Dataset}}     & \multirow{2}*{\textbf{Description}}   & \multirow{2}*{\textbf{Subset}} & \multirow{2}*{\textbf{Size}}  & \multirow{2}*{\shortstack{\textbf{Answer}\\\textbf{Type}}}\\ 
     & & & & \\
    \midrule
    StrategyQA~\citep{strategyqa} & QA & train & 2290 & T/F\\ 
    AQuA \citep{ling2017program}    & Algebra QA & test & 254 & Numbers\\ 
    Creak\citep{onoe2021creak}     & Commonsense Reasoning & dev & 1371  & T/F\\ 
    NoahQA\citep{zhang2021noahqa}     & Numerical Reasoning QA  & test  & 10880  & Multi\\ 
    GSM8k\citep{cobbe2021gsm8k}     & Math Reasoning   & test & 6140   & Text\\ 
    bAbi15\citep{dodge2016evaluating}     & Deductive Reasoning   & test & 1000    & Word\\ 
    bAbi16\citep{gaunt2017ampnet}     & Inductive Reasoning   & test & 1000    & Word\\ 
    QASC\citep{khot2020qasc}        & Multi-hop Reasoning       & dev & 926        & Word\\ 
    ECQA\citep{aggarwal2021explanations}        & Commonsense Reasoning      & test & 2194        & Word\\ 
    e-SNLI\citep{camburu2018snli}        & Logical Relationship Reasoning      & test & 9824        & Word\\ 
    Sen-Making~\citep{wang2019does}       & Commonsense Reasoning      & test & 2020        & Text\\ 
    QED~\citep{lamm2021qed}        & QA with explanations      & dev & 1354        & Text\\ 
    \midrule
     Total        & & & 39253        &\\ 
    \bottomrule
  \end{tabular}
  \end{small}
\end{table}

\subsection{Data Assembly}
\label{Data Assembly}
As we mentioned, we ran overall more than a million queries on the chosen benchmarking LLMs.
To cope with the scaled number of query responses, a must-do preliminary step is certainly to figure out an \emph{automatic interpreter} that manages to assess each individual response with no explicit requirement of human scrutiny or intervention.

That said, as a core component for the automatic interpreter, the raw form drawn from the original dataset requires amendment.
To that regard, we design a template with a uniform data form that inherent is a \emph{multiple-option} QA \textbf{data primitive}, showed in Table~\ref{tab:structure} where we demonstrate a set of samples.
In particular, we represent the target template as quintuplet form \data.
Here, we write down a set of confusing options that (i)-are not the correct option but (ii)-but similar to the ground-truth answer $a$.
The parameter $prompt$ denotes the prompts. Notably, $a$ won't be particularly long paragraphs and ChatGPT is able to output in our expected way under our setting of $o$.
Essentially, this templated data form urges the LLM to correctly differentiate the correct answer $a$ against the others in $o$, based on the passage $\mathbf{p}$ and the connected question $q$.
Thanks to the nature of this multiple-option form, we may simply scan the responses spit from the LLM for the index (e.g. A, B, C, D...) and pave the way for accurate calculation.
Guided by good prompts $\mathbf{p}$, this auto-conversion mechanism empirically works well with very few exceptions.

In the implementation, we first build a converter to automatically fetch $(\mathbf{p},q,a)$ part except for confusion options. For most datasets(as Appendix~\ref{appendix:datasets} shows) that provide passages and questions separately, we extract corresponding fields directly. Taking the example in Table~\ref{tab:structure}, contexts of `facts', `question', and `answer' fields correspond to $(\mathbf{p},q,a)$ separately.

This above mechanism shall already suffice for the datasets that expectedly consist of a passage.
For those that do not contain the passages, we alter the primitive by the following steps: (i)- we divide the question into two parts where the prior part involves and only involves the statement or the conditions, while the latter part is reduced to be the question. The process is automatically implemented by simple identification and sentence segmentation; (ii)-for the datasets where the questions are not decomposable, we simply set $\mathbf{p}$ to Null.

\paragraph{Confusion Option Generation}
We devise a generator that is dedicated to producing some confusion options $o$ according to answers $a$. 
As the core component of the primitive, the LLM is expected to choose $a$ against these confusion counterparts.
From Table~\ref{Dataset}, we may summarize that the answers in these datasets primarily are categorized by the following four types: T/F(True or False), Number, Word, and Text. 
For $o$ in this primitive, a good confusion option ought to be maximally similar to the ground truth but can be differentiated by the common-sense understanding or reasoning capacity of the LLM. 
Samples of confusion options are shown in Table ~\ref{tab::options generation}
In particular, for the distinct four categories, we devise the generated as follows:
\begin{itemize}
    \item \textbf{For datasets already provide a number of wrong options} (e.g., AQuA, QASC, ECQA, and e-SNLI), we directly use these options as $o$.
    \item \textbf{For T/F binary answers}, e.g., StrategyQA and Creak. While the built-in form of T/F should suffice for our primitive, we additionally incorporate a third option in $o$ as \texttt{unable to determine} to increase the difficulty.
    \item \textbf{For Numbered answers}, e.g., NoahQA(part of queries), we randomize four numerical values to form $o$. Hence, the total option number is 5.
    \item \textbf{For Word-ed answers}, e.g., bAbi15, bAbi16, ECQA. We randomly select the other words of the same linguistic property, and upon the results, we run from the Part-of-Speech tagger, \citet{1995Transformation}, in passages and/or questions. The total number in $o$ for this case is also 5.
    \item \textbf{For Text-ed answers}, e.g., GSM8K, . Basically, we alternate among the following three means: i)-delete, insert, or replace some words, ii)-change the correctness of the formula (if existed) or replace formulas of the current step with that of other steps and iii)-remove ground-truth $a$ and add an extra option `\texttt{None of the other options is correct}'. 
\end{itemize}

\begin{table}[htb]
  \caption{Confusion Options Generation Samples. For easier observation, we put all correct options in A-place. We actually randomize the order of the options as model's input.}
  \label{tab::options generation}
  \centering
  \begin{small}
  \begin{tabular}{c|p{0.2\linewidth}|p{0.6\linewidth}}
    \toprule
    \multirow{2}*{\shortstack{\textbf{Answer}\\\textbf{Type}}}     & \multirow{2}*{\shortstack{\textbf{Original}\\\textbf{Answer}}}   & \multirow{2}*{\shortstack{\textbf{Generated Options}}} \\ 
     & & \\
    \midrule
    T/F & True & \makecell[l]{\textbf{(A) True} \\ (B) False \\ (C) Unable to determine}\\ 
    \midrule
    Numbers & 36 & \makecell[l]{\textbf{(A) 36} \\ (B) 15 \\ (C) 17 \\ (D) 5 \\ (E) 7}\\ 
    \midrule
    Word & discovery & \makecell[l]{\textbf{(A) discovery} \\ (B) action \\ (C) reflection \\ (D) deciding \\ (E) thinking}\\
    \midrule
    Text & Janet sells 16-3-4=9 duck eggs a day & \makecell[l]{\textbf{(A) None of the other options is correct.} \\ (B) Janet elude sells 16-3-4=9 duck eggs a axerophthol day. \\ (C) Janet sells 4-4-10=-10 duck eggs a day. \\ (D) Janet sells 11-11-15=28 duck eggs a day. \\ (E) He has to pay 3000-1000=2000.}\\ 
    
    \bottomrule
  \end{tabular}
  \end{small}
\end{table}

\begin{table}[htb]
  \caption{The construction of model input from the original sample. The sample in this case was taken from the dataset StrategyQA. Origin structure shows raw data form. And the target data form, divided into some parts, is shown in other structure rows in order.}
  \label{tab:structure}
  \centering
  \begin{small}
  \begin{tabular}{l|p{0.8\linewidth}}
    \toprule
    \textbf{Structure} & \textbf{Sample} \\
    \midrule
    Origin & \makecell[l]{question\textquotedblright: ``Is Antarctica a good location for Groundhog Day?\textquotedblright,\\
        ``answer\textquotedblright: false,\\
        ``facts\textquotedblright: [\\
            ``Groundhog Day relies on a groundhog seeing their shadow.\textquotedblright,\\
            ``Antarctica has an irregular sun pattern and some days have no sun rise or 24 \\
            hour sunlight.\textquotedblright,\\
            ``Antarctica has temperatures can range from -10C to -60C.\textquotedblright,\\
            ``Groundhogs live in forests or woodlands with plenty of sunlight.\textquotedblright\\
        ]}\\
    \midrule
    \midrule
    Prompt &  Next, I will ask you a series of questions given a description, and you will have to choose one of several candidate options that you think is correct.  The description is\\
    \midrule
    Paragraph & Groundhog Day relies on a groundhog seeing their shadow.Antarctica has an irregular sun pattern and some days have no sun rise or 24 hour sunlight.Antarctica has temperatures can range from -10C to -60C.Groundhogs live in forests or woodlands with plenty of sunlight.\\
    \midrule
    Question & Is Antarctica a good location for Groundhog Day? \\
    \midrule
    Options & \makecell[l]{(A) True \\ \textbf{(B) False} \\ (C) Unable to determine} \\
    \midrule
    \midrule
    Query & \makecell[l]{Query 1: \{\emph{Prompt}\}+\{\emph{Paragraph}\} \\ 
            Query 2: The first question is \{\emph{Question}\}, choose an answer from the following\\ options: \{\emph{Options}\}.
            }\\
    \midrule
    Response & \textcolor{red}{The answer would be (B) False}. Antarctica is not a good location for Groundhog Day since the area has an irregular sun pattern, and some days could have no sun rise, or 24-hour sunlight. Groundhogs rely on seeing their shadow to predict the weather, and the irregular patterns of sunrise and sunset would make the whole concept unworkable in Antarctica. \\
    \bottomrule
  \end{tabular}
  \end{small}
\end{table}

\paragraph{Prompt Construction and In-context Learning}
We manually construct a fixed template as $prompt$, whose format is:
\begin{spacing}{1}
\textit{Next, I will ask you a series of questions given a description, and you will have to choose one of several candidate options that you think is correct. The description is}.
\end{spacing}
Based on this $prompt$ template and In-context Learning(ICL, \citet{brown2020language}), we divide a sample \data into queries. And we give queries to the model step by step in the same chat process, serving as In-Context Learning of ChatGPT. As the `Query' row in Table ~\ref{tab:structure} shows, settings of queries contain i) A context directly connected by $prompt$ and passage $\mathbf{p}$; ii) A context composed by question $q$, options $o$ and other linking clauses that make sentences flow. iii) If a sample provides more than one question for a passage, we will form follow-up queries with each question and corresponding options in turn.
After our verification, this template and ICL setting are (almost) universally feasible for constraining ChatGPT to answer the questions provided in our expected way.

We follow with evaluation methods surrounding robustness, consistency, and credibility. By this means, we hope to provide a throughout reference for ChatGPT's potential risks against real-world applications. Before that, we give some setups. For each dataset $D$, it is denoted as:
\begin{eqnarray}
\label{formula:1}
\begin{array}{c}
\displaystyle D=\{\mathbf{x}_i\}_{i=1}^n=\{(\mathbf{prompt}_i,\mathbf{p}_i,q_i,o_i,a_i)\}_{i=1}^n
\end{array}
\end{eqnarray}
where $\mathbf{x}_i$ denotes i-th sample in $D$, and the definition of \data is showed in Section ~\ref{Data Assembly}.

\subsection{Gauge the Robustness}
\label{subsection_gaugerob}
The upfront property of the LLM we intend to gauge is its robustness.
In a nutshell, we draw inspiration from the conventional NLP community --- in particular, those research centered around adversarial examples on text.
Borrowing the terms from \citet{zhang2022interpreting}, a perturbed example input is  defined as
\begin{eqnarray}
\label{formula:2}
\begin{array}{c}
\displaystyle x'=x+\delta {;} \parallel\delta\parallel \le \epsilon \land f(x,\mathbf{\theta}) \ne f(x,\mathbf{\theta}) 
\end{array}
\end{eqnarray}
where $\delta$ denotes the perturbation inserted into original sample $\mathbf{x}$ yielding its perturbed counterpart $\mathbf{x}'$. $f(x,\mathbf{\theta})$ represents the model in a functional form parametrized by $\mathbf{\theta}$.

Generally, in the conventional robustness assessment of NLP models, the $\epsilon$ is expected to be extremely small.
In spite of that, we shift the extensive emphasis on minimizing the distance/difference from the adversarial examples to their clean counterparts.
By contrast, we stress constructing the adversarial examples via more \emph{structural} means that are more compatible with the realistic deployment of the LLMs, rather than random adversarial perturbation.
For instance, with the release of the ChatGPT APIs, an (exponentially) increasing number of user-generated inputs would be fed to the API, followed by the distribution of the LLM's responses.
In this work, we delve into it and gauge mistakes --- that are common in user-generated content --- such as character-level typos (e.g typing error), incorrect words (e.g. user input obtained from speech transcription), or visual flaws (e.g. user input yielded from OCR), etc.
We believe this setup may simulate the possible outcomes for the particular deployment of the LLMs, being a better simulation than the study around conventional adversarial examples.

We carry out an adversarial attack(constructing adversarial examples) process with an auto-attacker $g$. Through this attacker, we can transfer the process to any dataset, as a part of the general automated evaluation system. Following conventional setting~\citep{wang2018glue, wang2021adversarial, nie2019adversarial}, we only perform adversarial attacks on $\mathbf{p}$ if $\mathbf{p}$ exists($q$, otherwise). The result in a sample $\mathbf{x}$ set adversarial example $\mathbf{x}'=(P,\mathbf{p}',q,o,a)$, and $\mathbf{p}'$ and $\mathbf{p}$ is denoted as:
\begin{equation}
    \begin{array}{c}
    \mathbf{p}'=[\mathbf{w}_{1}^{*},\cdots,\mathbf{w}_{n}^{*}]\\
    \mathbf{p}=[\mathbf{w}_{1}^{},\cdots,\mathbf{w}_{n}^{}]
    \end{array}
\end{equation}
where $w$ or $\mathbf{w}^*$ denotes word in $\mathbf{p}$ or $\mathbf{p}'$, and $\mathbf{w}^*$ can be further denoted as:
    \begin{equation}
    \label{formula:3}
    \begin{array}{c}
    z \sim \mathcal{U}(0,1) \\
    \mathbf{w}_i^*=\left\{\begin{array}{ll}
    \mathcal{g}(\mathbf{w}_i) & 0< z < \rho, \\
    \mathbf{w}_i & Otherwise
    \end{array}\right.
    \end{array}
    \end{equation}
where $\mathcal{U}$ is a uniform distribution, and $\rho$ is a preset probability between $[0,1]$, representing the probability of a word to be attacked.

Now, we go to the specific instantiations of auto-attacking function $g$ on each word $w$ from passage $\mathbf{p}$.
Noted again, we want to simulate the specific usage of LLMs.
The robustness of the considered LLM for the following setup is gauged by the calculation of the accuracy alter by the perturbed word against the original ones.
With the examples provided in Table~\ref{tab:sample}, we define the following methods to complement the auto-attacking function $g$. Notably, all the attacks focus on each word $w$. While each $w$ maintains a probability $\rho$, mentioned above, of being altered.
\begin{itemize}
\label{attack method}
    \item \textbf{Word-level attacks.} 
    For each $w$ that should be altered, we randomly conduct one of the three operations(i.e., insertion, deletion, and replacement). 
    Notably, for replacement and insertion, altered word $\mathbf{w}^*$ has a 50\% chance to be a random word from the original passage $\mathbf{p}$.
    Otherwise, $\mathbf{w}^*$ are synonyms of $w$ or random words from WordNet\citet{miller1995wordnet}.
    Word-level attack simulates word errors in applications, e.g., missing or misidentifying words in speech recognition.
    \item \textbf{Character-level attacks.} We randomly select characters of $w$, to be altered, according to a preset proportion. Subsequently, we perform one of the operations(i.e.,  repeat, insert, and delete) on selected characters. The settings of the proportion and inserted characters are listed in Appendix~\ref{appendix:attack}. Character attack simulates those human natural errors, like typing or spelling mistakes, in user-generated contexts.
    \item \textbf{Visual-level attacks.} Similar to character attacks, We select English letters in $w$, to be altered, by a preset ratio(see Appendix~\ref{appendix:attack} for details). And we perform substitution of these letters with visual-similar characters. The data source of these characters is derived from previous work~\citep{eger2019text}. This attack simulates visual flaws in some real-world applications, e.g., OCR.
\end{itemize}

\begin{table}[htb]
  \caption{Attack Description. The context inside the red brackets refers to the Unicode of the new character.}
  \label{tab:sample}
  \centering
  \begin{small}
  \begin{tabularx}{\textwidth}{X|l|p{0.65\linewidth}}
    \toprule
    Type & Specific & Sample\\
    \midrule
   Original & / & George Washington died in 1799.CDs weren't invented until 1982.\\
    \midrule
    \multirow{3}{*}{Character Level}
    & repeat & \textcolor{red}{Georggge} Washington \textcolor{red}{diiied iiin} \textcolor{red}{177799}.CDs weren't \textcolor{red}{innnvented} until 199982.  \\
    \cmidrule(r){2-3}
     & delete &  George Washington died in 1799.CDs weren't \textcolor{red}{inted un} 1982.    \\
    \cmidrule(r){2-3}
    & insert & George Washington \textcolor{red}{di@ed} in \textcolor{red}{1799@}.CDs weren't invented until 1982. \\
    \midrule
    \multirow{3}{*}{Word Level}  & insert &  \textcolor{red}{died} George Washington died in 1799.CDs \textcolor{red}{1799.CDs} weren't invented \textcolor{red}{hoosier state} until 1982. \\
    \cmidrule(r){2-3}
    & delete &  invented.    \\
    \cmidrule(r){2-3}
    & replace & George \textcolor{red}{cook up cook up 1982} 1799.CDs \textcolor{red}{go bad} invented until 1982. \\
    \midrule
    Visual Attack & / & 
    {George Washin\textcolor{red}{\{0260\}}ton \textcolor{red}{\{0257\}}ied in 1799.CDs weren't \textcolor{red}{\{0269\}}nvented unti\textcolor{red}{\{0625\}} 1982. }
    \\
    \bottomrule
  \end{tabularx}
  \end{small}
\end{table}

\subsection{Gauge the Consistency}
As we mentioned, we use the term \emph{consistency} to measure the consistency of the LLMs when faced with user-generated inputs that convey similar semantics.
Thereby, we conduct design two novel attacking methods that reflect two distinct aspects.

\paragraph{Prompt Attack}
In real-world applications, different users may have diverse expressions for the input of the same meaning. These inputs are semantically similar but in very different forms. It is of great importance for ChatGPT to maintain consistency against these inputs.
To conduct the simulation, we select five statements, denoted as $prompt'$, synonymous with $prompt$(listed in Table~\ref{tab:prompt_sample}). 
And we alternately replace $prompt$ with different $prompt'$, creating an automated process.
In this process, we test ChatGPT's output against different $prompt'$, to measure consistency for different prompts.
\begin{table}[htb]
\caption{Prompts for the multiple-option test.}
  \centering
  \begin{tabularx}{\textwidth}{p{0.7\linewidth}|X}
    \toprule
    \textbf{Text} & \textbf{Source} \\
    \midrule
     ``\textquotedblright& Blank prompt \\
     \midrule
    ``Complete the description with an appropriate ending: \textquotedblright & Valid prompt of OPT\\
    \midrule
    ``You must choose the best answer from the following choices marked (A), (B), (C), (D) or (E).\textquotedblright & \multirow{2}{*}{CET examination}\\ 
    \midrule
    ``To answer the following question according to the following information.\textquotedblright & \multirow{2}{*}{Human-made} \\
    \midrule
    ``Next, I will ask you a series of questions given a description, and you will have to choose one of several candidate options that you think is correct.\textquotedblright &  \multirow{3}{*}{Human-made} \\
    \bottomrule
  \end{tabularx}
  \label{tab:prompt_sample}
\end{table}
\paragraph{Options Attack}
For the other, we randomize the order of options --- $a$ and $o$ combined --- in  Formula~\ref{formula:1} in order to test LLMs' performance.
It is apparent that this has no impact on the semantics of the data primitive itself, but empirically we find certain LLM exhibit weird behavior against this attack that we show in the experiments section. 
This attack method simulates some parallel grammatical structure, whose order does not affect reading, in actual usage. 

\subsection{Gauge the Credibility: Relative Training Index.}
\label{subsection:rti}

Indeed, this \emph{credibility} gauging is admittedly a side-product of the empirical study of this work.
That is when we conduct the aforementioned attack method in extreme strength and magnitude, LLMs are still able to perform accurately.
Similar to the VQA benchmarking issue(\citet{10.1007/978-3-319-46484-8_44}), this cast serious concerns on the current benchmarking methods, especially when being used to assess LLM-involved methods.
Thus, in spite of being a side-product, the importance of this section shall not be deprioritized nor overlooked.
In particular, we present a novel indexing number --- dubbed as Relative Training Index,  --- that measures the \emph{relative} reliability of the provided datasets towards LLM-related evaluation.

Formally, the \rti{} is calculated by the following steps (specific complementation is showed in Algorithm ~\ref{alg::rti})
\begin{itemize}
    \item We adopt the word-level attack due to its commonality. As per Formula~\ref{formula:3}, we gradually increase the parameter $\rho$ from $0$ to $1$ by a stride of 0.1. Within each strived point, we test the LLM towards its final accuracy.
    \item We detect the output of the model for each $\rho$. During this process, we find the minimum $\rho$, denoted as $r$, that causes the model to change the original answer.
    \item Finally, we calculate the expectation of $r$ for each $\mathbf{x}$ in dataset $D$ as the $R_D$(\rti{}). \rti{} can be denoted as:
    \begin{eqnarray}
    \begin{array}{c}
    \displaystyle \mathbf{R}_D=\mathbb{E}_{\mathbf{x} \in D} r_\mathbf{x} \\
    \end{array}
    \end{eqnarray}
\end{itemize}

\begin{algorithm}[htb]
  \caption{Calculate \rti{} $\mathbf{R}_D$ of dataset $D$}
  \label{alg::rti}
  \begin{algorithmic}
    \Require
    \\
      $D=\{\mathbf{x}_1,\mathbf{x}_2,\cdots,\mathbf{x}_n\}$ : dataset\\
      $g(x,\rho)$ : auto-attacker on $\mathbf{x}$ with probability $\rho$\\
      $f(x,\mathbf{\theta})$ : model output on $\mathbf{x}$ with parameter $\mathbf{\theta}$
    \Ensure
    \\
      $\mathbf{R}_D$ : \rti{} score of dataset $D$
    \end{algorithmic}
    
   \begin{algorithmic}[1]
    \For {$\mathbf{x}_i$ in $D$}          
        \State $\rho = 0.1$
        \State $\mathbf{x}_i' = g(\mathbf{x}_i,\rho)$ 
        \While {$f(\mathbf{x}_i,\mathbf{\theta}) = f(\mathbf{x}_i',\mathbf{\theta})$}
            \State $\rho = \rho+0.1$
            \State $\mathbf{x}_i' = g(\mathbf{x}_i,\rho)$
        \EndWhile
        \State $r(\mathbf{x}_i)=\rho$
    \EndFor
    \State $\mathbf{R}_D=\mathbb{E}(R)$ where $R \sim r(\mathbf{x}),\mathbf{x} \in D$  
    \State return $\mathbf{R}_D$;
  \end{algorithmic}
\end{algorithm}

We intend to assess the reliability of the normal NLP task evaluation that particularly involves an LLM.
Intuitively, if the dataset was chosen to fill in the overall training set for the particular LLM, by proper training flow, the LLM should have absorbed and memorized the aspects of this dataset.
In that case, incrementally developing an external module to couple the LLM for this dataset --- or its involved task --- would be somewhat problematic.
Put another way, we find that \textbf{in some occasions, LLM can respond correctly even when the passage sample is 100\% polluted!} 
This would evidently mean that the LLMs have already encoded the information and/or its answer into their parametric memorization. Hence, further evaluation of LLM-involved tasks on such occasions would very much be misleading.
Further, we would admit that this study is just an implication of the reliability problem rather than fully explaining or resolving the puzzle --- due to that, there is not a transparent path towards reverse-engineering on the original training set for such LLM (especially ChatGPT).

Put another way, indicated by $\rti{}$, the higher the score the lower reliability of the developed method is, and vice versa.

\textbf{(Perhaps wild) claim.}
In hindsight, lower $\rti{}$ shall encourage our cohorts to move away from the considered datasets in the era of the LLM.

\subsection{Metrics}
\label{sec:metric}
\label{Robustness and Stabiliry Test}
\subsubsection{Metrics on Robustness and Consistency.} 
To finalize the gauging process on the popular LLMs, we devise two complementary quantitative measurements: the error rate($\mathbf{ER}, \%$) and the answer-changing Rate($\mathbf{ACR}, \%$).
In particular, the error rate resembles conventional NLP evaluation where we judge the LLM based on its accuracy in correctly responding to the data primitive.
On the other hand, the error-changing rate measures the proportion from the testing set that the model maintains the original output post to the attacks.

We write down the formulation of the two metrics as follows:
\begin{equation}
\mathbf{ER}_D = \frac{\mid\{\mathbf{x}_i\mid \mathbf{x}_i \in D\land f(\mathbf{x}_i,\mathbf{\theta}) \ne a_i\}\mid}{\mid D \mid}
\end{equation}
\begin{equation}
\mathbf{ACR}_D = \frac{\mid\{\mathbf{x}_i\mid \mathbf{x}_i \in D\land f(\mathbf{x}_i,\mathbf{\theta}) \ne f(\mathbf{x}_i',\mathbf{\theta})\}\mid}{\mid D \mid}
\end{equation}
where $\mid  \mid$ denotes a counter.

\subsection{Qualitative Pattern Analysis of Adversarial Sample}
Besides the quantitative measurements, we intend to thoroughly analyze the LLMs' behavior against our primitive.
In that, we next introduce our qualitative pattern analysis scheme.
We mostly follow previous work including ~\citet{nguyen2019identifying, goodman2020fastwordbug, zheng-etal-2020-evaluating, xue2020dpaeg, wang2021towards}.
Likewise, upon gathering the LLMs' responses from the corresponding input --- that includes both the clean and polluted counterpart --- we aim to analyze which part/component of the input prompt may maximally and most probably cause the LLM to drift.

In that regard, we leverage tools like past-of-speech taggers, dependency parsing, phrase structure discovery\footnote{https://github.com/stanfordnlp/CoreNLP} \footnote{https://github.com/explosion/spaCy}, and intra-sentence positional analysis.
We ground this portion of the analysis on the granularity of words, which covers all the attacking means as we mentioned before.

More specifically, we devise a separate mechanism for the above sources of information:
\begin{itemize}
    \item For structure and intra-sentence analysis, we consider attacked samples $\mathbf{x}'$ with unexpected output, where $f(\mathbf{x}',\mathbf{\theta}) \ne f(\mathbf{x},\mathbf{\theta})$. We count different categories' occurrence frequency of perturbed parts in $\mathbf{x}'$. Categories $l$ here are position tags(e.g., head, tail, or middle positions) or structure tags(e.g., Noun Phrase, Noun Phrase). The frequency, denoted as $\mathbf{s}_l$, can be represented as:
    \begin{equation}
    \label{formula:9}
        \begin{array}{c}
        \displaystyle \mathbf{s}_l = \sum_{\mathbf{x} \in D} \frac{1}{G(\mathbf{x})} \sum_{\mathbf{w} \in \mathbf{x},\mathbf{w} \ne \mathbf{w}^*}\mathbf{sgn}(\mathbf{w}^*,\mathbf{x},\mathbf{x}',l) \\
        \displaystyle G(\mathbf{x}) = \mid \{\mathbf{w}_i\mid \mathbf{w}_i \in \mathbf{x} \land \ \mathbf{w}_i \ne \mathbf{w}^*_i\}\mid \\
        \mathbf{sgn}(\mathbf{w}^*,\mathbf{x},\mathbf{x}',l)=\left\{\begin{array}{ll}
        1 & l(\mathbf{w}^*) = l \land f(\mathbf{x},\mathbf{\theta}) \ne f(\mathbf{x}',\mathbf{\theta}) \\
        0 & otherwise
        \end{array}\right.
        \end{array}
    \end{equation}
    where $l(\mathbf{w})$ is a function to get the category of the word $\mathbf{w}$, $\mathbf{sgn}$ is a function measuring whether the sample cause the LLM to drift, $G$ is a regularization function to balance attack times.
    \item And for past-of-speech and dependency parsing analysis, similarly, we evaluate them by calculating their contributions to the successful attacks.
    Similar to the above, we firstly formulate a vector $\mathbf{c}$ by counting the normalized frequency of each $l$ of categories set $L$:
    \begin{equation}
    \label{formula:8}
    \mathbf{c} =
        \left[ 
            \frac{\mid \{ \mathbf{w}\mid \forall \mathbf{w} \in \mathbf{x}, \ \mathbf{w}^* \ne \mathbf{w} \land l(\mathbf{w})=l_j \mid}
            {\mid \{ \mathbf{w}\mid \forall \mathbf{w} \in \mathbf{x}, \ l(\mathbf{w})=l_j \} \mid } 
        \right]_{l_j \in L}
    \end{equation}
    Then, by checking every attacked sample in $D$, we can extract a set $\{(\mathbf{c}_i, sgn(\mathbf{x}_i, \mathbf{x}'_{i}))\}_{i=1}^{n}$, where $sgn(\mathbf{x}_i, \mathbf{x}'_{i})$ returns 1 if $f(\mathbf{x},\mathbf{\theta}) \ne f(\mathbf{x}',\mathbf{\theta})$, and 0 otherwise.
    Secondly, we train a Random Forest (RF) on this set, with $\mathbf{c}_i$ and $ \mathbf{sgn}(\mathbf{x}_i, \mathbf{x}'_{i})$ as input feature and output target for training, respectively.
    Finally, we employ the trained RF model to assign an \emph{importance score} to each category related to the success of attacks.
\end{itemize}

\section{Experiment}

\subsection{Robustness Results}

\subsubsection{Main Results}
The robustness test results across adversarial attacks are listed in Table~\ref{tab:level attack}. And specific results of each dataset are listed in Table~\ref{rob_full_er} and Table~\ref{rob_full_acr}.

\begin{table}[htb]
  \caption{Different level attack influence on ChatGPT. (The results are tested on all datasets with 39253 samples in total.)}
  \label{tab:level attack}
  \centering
  \centerline{\begin{tabular}{c|c|cccc|ccc}
    \toprule
    Model & Attack Type  & \multicolumn{4}{c}{Error Rate(\%)} & \multicolumn{3}{c}{Answer Changed Rate(\%)}              \\
    \cmidrule(r){1-9}
    \multirow{6}{*}{ChatGPT} & \multirow{2}{*}{Character Level} & ori & repeat & delete & insert & repeat & delete & insert \\
    &  & 40.19 & 44.59 & 51.89 & 41.63 & 30.51 & 38.56 & 27.45  \\
    \cmidrule(r){2-9}
    & \multirow{2}{*}{Word Level}  & ori & insert & delete & replace & insert & delete & replace \\
    &  & 40.19 & 48.04 & 61.89 & 60.85 & 34.27 & 49.8 & 48.18  \\
    \cmidrule(r){2-9}
    & \multirow{2}{*}{Visual-level}  & ori & 10\% & 50\% & 90\% & 10\% & 50\% & 90\% \\
     & & 40.19 & 42.47 & 48.85 & 55.88 & 28.19 & 34.64 & 41.94 \\
    \bottomrule
  \end{tabular}}
\end{table}

\begin{table}[htb]
  \caption{Different level attack influence of each dataset on ChatGPT, measured by error rate(\%).}
  \label{rob_full_er}
  \centering
  \begin{small}
  \centerline{\begin{tabular}{lc|ccc|ccc|ccc}
    \toprule
    \multicolumn{2}{c}{Dataset}  & \multicolumn{3}{c}{Charater Level} & \multicolumn{3}{c}{Word Level} & \multicolumn{3}{c}{Visual Level}              \\
    \cmidrule(lr){1-2} \cmidrule(lr){3-5} \cmidrule(lr){6-8} \cmidrule(lr){9-11}
    Name    & ori  
    & repeat & delete & insert 
    & insert & delete & replace 
    &  10\% & 50\% & 90\% \\
    \midrule
    StrategyQA & 29.56 & 31.10 & \underline{37.46} & 31.09 & 35.48 & 46.33 & \pmb{\underline{46.77}} & 31.07 & 35.92 & \underline{43.65} \\
AQuA & 47.64 & 63.74 & \underline{74.22} & 51.40 & 68.31 & \pmb{\underline{89.57}} & 89.17 & 54.92 & 58.07 & \underline{71.26} \\
Creak & 34.14 & 34.31 & \underline{35.85} & 35.64 & 34.46 & 36.51 & \underline{36.98} & 36.40 & 38.04 & \pmb{\underline{39.75}} \\
NoahQA & 33.01 & 41.75 & \underline{47.83} & 35.41 & 40.51 & \pmb{\underline{63.10}} & 61.57 & 34.38 & 37.87 & \underline{46.68} \\
GSM8K & 54.8 & \underline{62.28} & 60.29 & 55.33 & 57.72 & \pmb{\underline{63.00}} & 62.15 & 54.57 & 55.50 & \underline{59.61} \\
bAbi15 & 29.0 & 28.21 & \underline{49.15} & 26.68 & 52.35 & \pmb{\underline{64.50}} & 64.30 & 31.35 & 45.15 & \underline{59.35} \\
bAbi16 & 55.4 & 52.97 & \underline{62.15} & 54.76 & 60.20 & \underline{61.65} & 61.10 & 56.80 & 60.50 & \pmb{\underline{66.25}} \\
ECQA & 26.94 & 31.48 & \underline{54.92} & 30.91 & 49.02 & 72.11 & \pmb{\underline{74.68}} & 33.91 & 51.23 & \underline{66.27} \\
ESNLI & 52.99 & 53.58 & \underline{58.98} & 52.65 & 58.37 & \underline{64.01} & 63.85 & 54.99 & 61.27 & \pmb{\underline{64.50}} \\
QASC & 20.52 & 23.01 & \underline{53.07} & 25.70 & 38.55 & \underline{70.09} & 64.09 & 29.27 & 57.13 & \pmb{\underline{72.68}} \\
QED & 23.54 & 27.21 & \underline{45.52} & 28.98 & 37.79 & \pmb{\underline{58.67}} & 58.63 & 29.70 & 44.06 & \underline{56.20} \\
SenMaking & 19.94 & 23.42 & \underline{34.22} & 22.67 & 31.89 & \pmb{\underline{59.30}} & 50.12 & 26.18 & 41.17 & \underline{49.04} \\
    \bottomrule
  \end{tabular}}
  \end{small}
\end{table}

\begin{table}[htb]
  \caption{Different level attack influence of each dataset on ChatGPT, measured by answer changed rate(\%).}
  \label{rob_full_acr}
  \centering
  \begin{small}
  \centerline{\begin{tabular}{l|ccc|ccc|ccc}
    \toprule
    \multirow{2}{*}{Dataset}  & \multicolumn{3}{c}{Charater Level} & \multicolumn{3}{c}{Word Level} & \multicolumn{3}{c}{Visual Level}              \\
     \cmidrule(lr){2-4} \cmidrule(lr){5-7} \cmidrule(lr){8-10}   
    & repeat & delete & insert 
    & insert & delete & replace 
    &  10\% & 50\% & 90\% \\
    \midrule
    StrategyQA & 25.31 & \underline{31.72} & 24.39 & 30.09 & \pmb{\underline{42.77}} & 42.05 & 25.44 & 31.94 & \underline{40.41} \\
AQuA & 64.62 & \underline{73.65} & 56.70 & 70.28 & 82.87 & \pmb{\underline{84.65}} & 56.30 & 60.63 & \underline{69.69} \\
Creak & 15.80 & \underline{19.62} & 16.69 & 18.02 & \underline{23.30} & 22.79 & 16.48 & 20.64 & \pmb{\underline{23.89}} \\
NoahQA & 34.31 & \underline{40.65} & 28.21 & 33.03 & \pmb{\underline{57.35}} & 54.38 & 27.14 & 30.73 & \underline{39.81} \\
GSM8K & 35.66 & \underline{37.41} & 27.71 & 32.16 & \pmb{\underline{45.89}} & 42.89 & 27.78 & 31.38 & \underline{36.12} \\
bAbi15 & 36.57 & \underline{57.30} & 36.43 & 58.00 & \pmb{\underline{67.95}} & 67.30 & 40.40 & 51.15 & \underline{62.45} \\
bAbi16 & 54.16 & \underline{68.05} & 56.34 & 68.15 & 69.70 & \pmb{\underline{71.05}} & 58.15 & 61.40 & \underline{67.25} \\
ECQA & 16.44 & \underline{42.13} & 16.13 & 38.47 & 63.81 & \pmb{\underline{68.69}} & 20.31 & 38.58 & \underline{56.06} \\
ESNLI & 31.32 & \underline{35.87} & 31.30 & 34.92 & 38.59 & \pmb{\underline{38.77}} & 32.01 & 35.91 & \underline{38.24} \\
QASC & 12.72 & \underline{43.08} & 13.05 & 29.16 & \underline{62.47} & 57.34 & 17.12 & 46.81 & \pmb{\underline{65.01}} \\
QED & 16.56 & \underline{39.00} & 16.41 & 28.19 & \pmb{\underline{53.03}} & 52.25 & 20.37 & 36.86 & \underline{51.99} \\
SenMaking & 19.23 & \underline{30.25} & 18.85 & 28.77 & \pmb{\underline{55.34}} & 46.29 & 22.09 & 36.07 & \underline{44.29} \\
    \bottomrule
  \end{tabular}}
  \end{small}
\end{table}

\paragraph{ChatGPT show low robustness.}
All attack methods have a negative impact on ChatGPT. Specifically, regardless of attack types, it leads to at least a 2\% increase in the $\mathbf{ER}$ of ChatGPT's responses. Besides, as $\mathbf{ACR}$ results show, all attacks cause the model to change at least 27\% of the output option. Furthermore, \textbf{Robustness against character-level attacks is higher, and robustness against word-level attacks is lower among three levels}. Taking ER as an instance, the average $\mathbf{ER}$ on word-level attacks is roughly $10\%$ higher than the character-level. Simultaneously, $\mathbf{ER}$ on word-level attacks shows higher variance on different method types. Results in $\mathbf{ACR}$ indicator are similar. One possible reason is that word-level attacks caused larger changes in encoded vector by tokenizer\cite{10.1007/978-981-15-6198-6_18} and word-embedding\cite{almeida2023word}. As model input, this change makes a greater impact on the model's semantic understanding.

For each attack level, we now proceed to a detailed analysis.
\begin{itemize}[itemsep=1pt,topsep=0pt,parsep=0pt]
    \item \textbf{Character-level.} In this level, deletion is a more effective attack method, while insertion has less impact. $\mathbf{ER}$ on deletion is $10\%$ higher than insertion, while $\mathbf{ACR}$ is $11\%$ higher. But the difference is not significant between various methods at this level. Character-level attacks are likely to be introduced by human natural errors, e.g., spelling mistakes and typo errors. More potential risks in some application scenarios are discussed in Section ~\ref{discussion_and_conclusion}.
    \item \textbf{Word-level.} $\mathbf{ER}$ and $\mathbf{ACR}$ in word-level methods maintained at a relatively high level. Specifically, deletion operation has the strongest impact, with $61.8\%$ in $\mathbf{ER}$ and $49.8\%$ in $\mathbf{ACR}$. Robustness against word-level attacks may pose potential risks in real-world applications, e.g., speech recognition. In these scenarios, input texts can usually ensure each word's completeness. However, the correctness of words among contexts cannot be always guaranteed.
    \item \textbf{Visual-level.} The higher the proportion of transformations adopted on a single word is, the higher $\mathbf{ER}$ and $\mathbf{ACR}$ is. However, the model can still maintain around 40\% accuracy. This case may be due to the rewriting of GPT Tokenizer considering visually similar characters. Visual robustness is crucial for the real-world application of LLMs in OCR and other scenarios.
\end{itemize}

Also, we conduct comparative experiments on other LLMs (including LLaMA\cite{touvron2023llama}, OPT\cite{zhang2022opt}). We choose datasets $D$ for ChatGPT with the best (SenMaking) and worst (bAbi-task 16) performance in $\mathbf{ER}$ as test benchmarks. Detailed model information and experimental results are listed in Table~\ref{tab:different model ER} and Table~\ref{tab:different model ACR}. \textbf{Compared to other models, ChatGPT performs better on $\mathbf{ER}$ metric, but worse on $\mathbf{ACR}$ metric.} The possible reason for the former is that other models themselves have poor baseline abilities. 
For the latter, we found that some models have high randomness and low confidence in their output options. We can indicate that models have not fully comprehended the intention to make multi-option.

\begin{table}[htb]
  \caption{Attack influence on different LLMs, reported in error rate(\%)}
  \label{tab:different model ER}
  \centering
  \begin{small}
  \setlength\tabcolsep{3pt}
  \centerline{\begin{tabular}{c|cccc|cccc}
    \toprule
    \multicolumn{1}{c}{Model}  & \multicolumn{4}{c}{bAbi16} & \multicolumn{4}{c}{SenMaking}              \\
    \cmidrule(r){1-9}
    Name  & samples & ori & word\_replace & chr\_delete & samples & ori & word\_del & chr\_insert \\
    \midrule
    ChatGPT(175B*) & \multirow{3}{*}{1000} & \underline{\pmb{55.40}} & \underline{\pmb{61.10}} & 62.15 & \multirow{3}{*}{2021} & \underline{\pmb{19.94}} & \underline{\pmb{59.30}} & \underline{\pmb{22.67}}   \\
    OPT(1.3B) & & 90.00 & 89.45 & 88.87 & & 100.00 & 100.00 & 100.00 \\
    LLaMA(11.5B) & & 61.90 & 81.10 & \underline{\pmb{56.85}} & & 56.46 & 64.77 & 54.66   \\
    \bottomrule
  \end{tabular}}
  \end{small}
\end{table}

\begin{table}[htb]
  \caption{Attack influence on different LLMs, reported in answer changed rate(\%)}
  \label{tab:different model ACR}
  \centering
  \begin{small}
  \centerline{\begin{tabular}{c|ccc|ccc}
    \toprule
    \multicolumn{1}{c}{Model}  & \multicolumn{3}{c}{bAbi16} & \multicolumn{3}{c}{SenMaking}              \\
    \cmidrule(r){1-7}
    Name  & samples & word\_replace & chr\_delete & samples & word\_del & chr\_insert \\
    \midrule
    ChatGPT(175B*) & \multirow{3}{*}{1000} & 71.05 & 56.34 & \multirow{3}{*}{2021} & 55.34 & 18.85   \\
    OPT(1.3B) & & \underline{\pmb{47.50}} & \underline{\pmb{44.66}} &  & \underline{\pmb{0.00}} & \underline{\pmb{0.00}}  \\
    LLaMA(11.5B)  & & 62.75 & 63.74 &  & 34.66 & 27.13   \\
    \bottomrule
  \end{tabular}}
  \end{small}
\end{table}

\subsubsection{Attack Pattern Analysis}
\begin{figure}[htb]
    \centering
    \includegraphics[width=\linewidth]{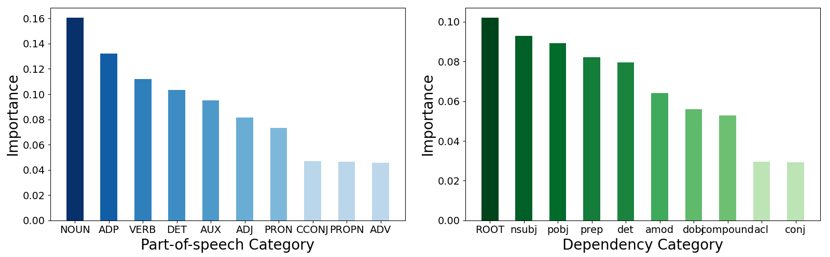}
    \caption{Attack pattern analysis on part-of-speech (left) and dependency relations (right). We evaluate the importance of different categories of attacked text for attack success through random forests. See text in Section~\ref{sec:metric} for details.}
    \label{fig:tag_dep}
\end{figure}

\textbf{From the perspective of part-of-speech (POS) and dependency relations: }Following the methodology outlined in Section~\ref{sec:metric}, we analyze the importance $c$(Formula ~\ref{formula:8}) of different POS tags and dependency tags of perturbed words. As shown in Figure~\ref{fig:tag_dep}, it indicates that adversarial attacks on ChatGPT can be easier to succeed when targeting specific POS-tagged or dependency-tagged words. Specifically, the top five tags that lead to successful attacks are $[\text{`NOUN', `ADP', `VERB', `DET', `AUX'}]$, and $[\text{`ROOT', `nsubj', `pobj', `prep', `det'}]$ (the concept illustration of these tags is showed in the website\footnote{https://github.com/clir/clearnlp-guidelines/blob/master/md/specifications/dependency\_labels.md} \footnote{https://github.com/explosion/spaCy/blob/b69d249a223fa4e633e11babc0830f3b68df57e2/spacy/glossary.py}), respectively. One potential reason for this could be the fact that these tags play a crucial role in determining the meaning and structure of a sentence. 
\begin{itemize}[itemsep=1pt,topsep=0pt,parsep=0pt]
    \item Nouns, verbs, and adpositions in particular are fundamental in expressing critical concepts and relationships within a sentence, while determiners and auxiliaries impact tense and agreement. 
    \item The `ROOT' dependency is the center word in a sentence that connects all other words and their dependencies. Similarly, `nsubj' and `pobj' dependencies are related to subject and object nouns respectively. The `prep' and `det' dependencies are related to prepositions and determiners respectively, which are crucial for defining and differentiating between objects and concepts. 
\end{itemize}

\begin{figure}[htb]
    \centering
    \includegraphics[width=\linewidth]{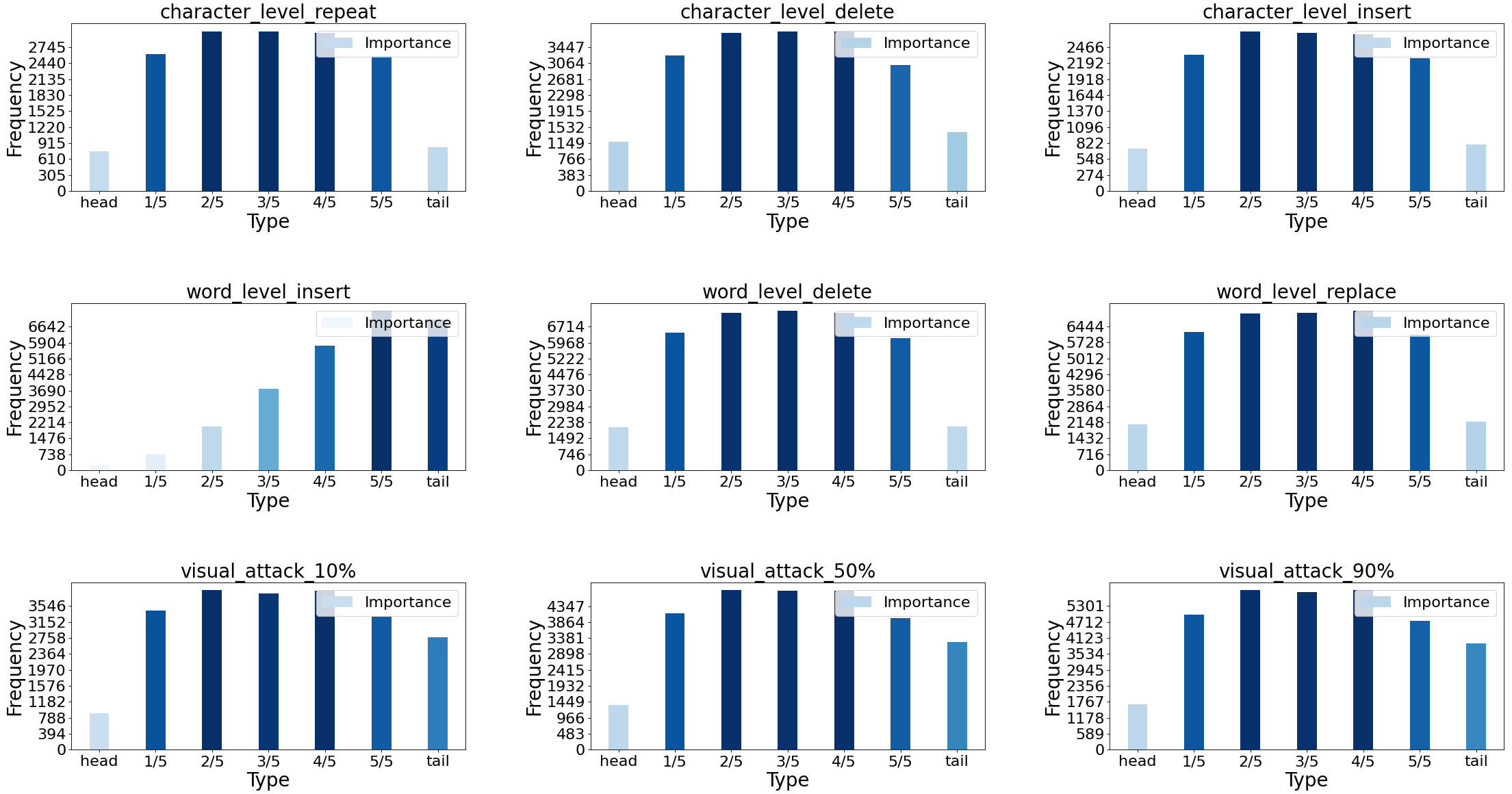}
    \caption{Attack pattern analysis on positions. We evaluate the importance of different positions (including head, tail, and relative positions of text) of attacked text for attack success through the frequency $s_l$(Formula ~\ref{formula:8}). See text in Section~\ref{Dataset and Method} for details.}
    \label{fig:postest}
\end{figure}

\textbf{From the perspective of intra-sentence position and phrase structure: }Following the methodology outlined in Section~\ref{sec:metric}, 
we get the result as Table~\ref{tab:parser} and Figure~\ref{fig:postest} shows. It is easier to change original responses when attacks occur in words with a certain category $l$ (Formula ~\ref{formula:8}). For parser structure tags, $[\text{`NP', `VP', `PP', `WHNP', `ADJP'}]$ (concept illustration is showed in the website\footnote{https://stanfordnlp.github.io/CoreNLP/parse.html\#description}) show the greatest impact. And for the position, the impact of attacking words in the middle is greater than attacking those at either end.
Now we will conduct a detailed analysis separately.
\begin{itemize}[itemsep=1pt,topsep=0pt,parsep=0pt]
    \item $l$ in [`Noun Phrase'(NP), `verb phrase'(VP), `preposition phrase'(PP)] shows a higher influence weight than any other categories on all datasets. Meanwhile, replacement at word-level is the most sensitive method to changes in the categories. One potential reason is that the language model generates the probability of the next word based on historical text. And nouns, verbs, and prepositional phrases can strongly affect the sentence structure. So that it will affect the prediction probability of correct words, and lead to a level of uncertainty in output. 
    \item The impact of attacking middle part words is significantly higher than attacking the head or tail words. The impact is evenly distributed within the middle parts and specific positions have little effect. And in some methods, attacking the end words may have a greater impact than attacking the head. Possible reasons for these situations are: \\
    \quad\quad1) A large number of words are involved in transitions in the middle. Since the length of the paragraph is relatively long.\\
    \quad\quad2) different positions have different attention weights within the model. 
\end{itemize}

Thus, perturbing these words may significantly impact the understanding of the original context or question. These patterns that are most susceptible to attacks can be identified. After that, future research can focus on developing more robust and safer LLMs to withstand these types of threats.

\begin{table}[htb]
  \caption{Parser Pattern Analysis}
  \label{tab:parser}
  \centering
  \setlength\tabcolsep{3pt}
  \small{\centerline{\begin{tabular}{l|cccccccccc}
    \toprule
    \multicolumn{1}{c}{Method}  & \multicolumn{10}{c}{Occurrence Frequency $s_l$} \\
    \cmidrule(r){1-11}
    Name     & NP & VP & PP & WHNP & ADJP & ADVP & QP & PRT & WHPP & WHADJP \\
    \midrule
char\_repeat & \pmb{2846.95} & 1382.41 & 1071.72 & 89.78 & 91.68 & 71.75 & 29.76 & 22.87 & 5.79 & 8.47 \\
char\_delete & \pmb{4297.89} & 1814.48 & 771.96 & 228.64 & 188.36 & 155.89 & 34.77 & 26.72 & 3.20 & 10.12 \\
char\_insert & \pmb{2598.74} & 1313.87 & 950.05 & 90.10 & 87.16 & 69.68 & 20.80 & 20.82 & 2.98 & 5.86 \\
word\_insert & \pmb{807.43} & 310.61 & 226.88 & 37.65 & 31.37 & 23.02 & 8.97 & 5.13 & 1.77 & 2.03 \\
word\_delete & \pmb{7452.55} & 3359.09 & 2216.4 & 373.21 & 221.78 & 215.27 & 64.14 & 42.37 & 16.31 & 13.46 \\
word\_replace & \underline{\pmb{8671.48}} & \underline{3537.06} & \underline{2372.13} & \underline{394.65} & \underline{284.26} & \underline{230.74} & \underline{76.84} & \underline{52.62} & \underline{16.84} & \underline{19.61} \\
visual\_10\% & \pmb{3629.25} & 2026.04 & 1447.34 & 135.76 & 138.66 & 108.13 & 31.71 & 33.69 & 6.22 & 7.56 \\
visual\_50\% & \pmb{4507.03} & 2483.14 & 1741.83 & 262.91 & 177.83 & 163.08 & 31.99 & 43.46 & 9.50 & 10.16 \\
visual\_90\% & \pmb{5512.11} & 3088.87 & 2085.59 & 358.46 & 226.42 & 210.19 & 36.28 & 50.31 & 12.24 & 13.43 \\
    \bottomrule
  \end{tabular}}}
\end{table}

\subsection{Consistency Results}
\begin{figure}[htb]
    \centering  
        \subfloat[The impact of prompts on accuracy]{  
        \begin{minipage}{0.95\textwidth}
        \centering    
        \includegraphics[width=\textwidth]{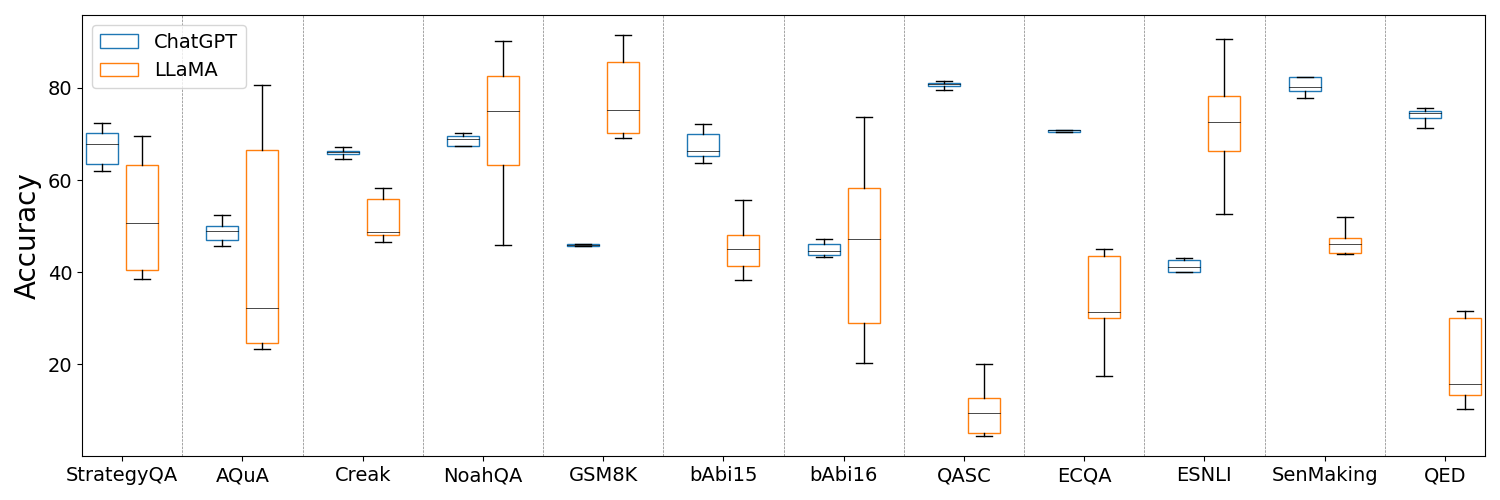}
        \end{minipage}
        }
        \\
        \subfloat[The impact of option orders on accuracy]{ 
        \begin{minipage}{0.95\textwidth}
        \centering    
        \includegraphics[width=\textwidth]{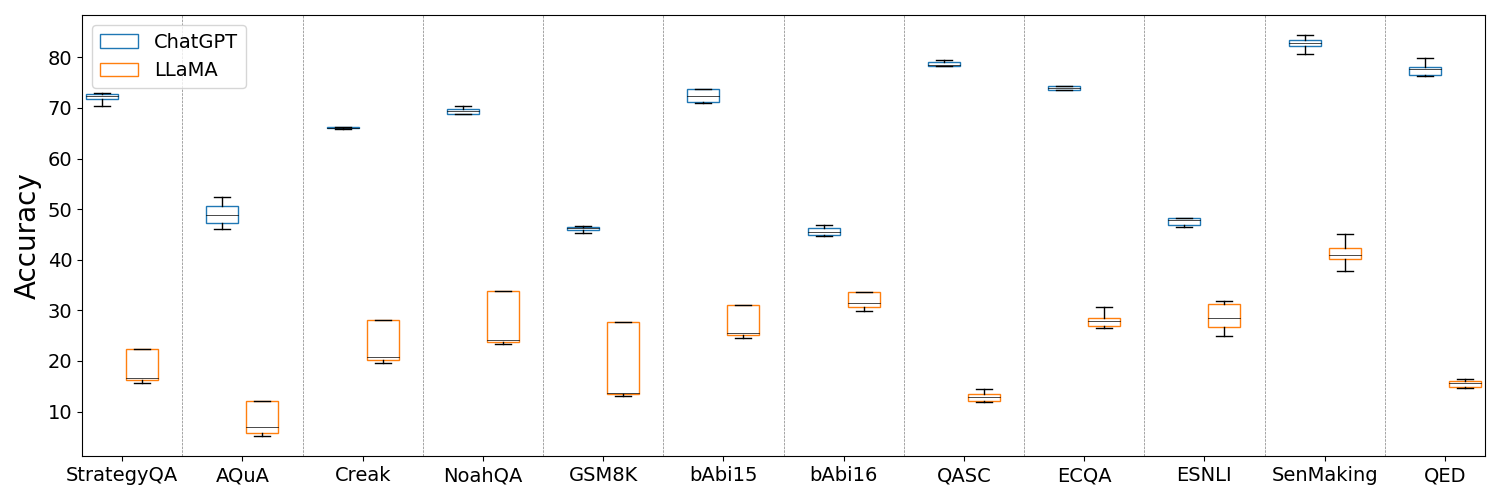}
        \end{minipage}
        }
    \caption{The impact of prompts and option orders on accuracy. For the former, the results are obtained by testing under 5 different prompts for the multiple-option test (as listed in Table~\ref{tab:prompt_sample}). While for the latter, the results are obtained by testing on 6 different option orders.}
    \label{fig:position}
\end{figure}

Figure~\ref{fig:position} present the consistency of ChatGPT across multiple runs on different $o$ and $a$ orders and $prompt$(Formula ~\ref{formula:1}). The results are reported using box plots, and we use the standard deviation of accuracy to represent the consistency of the results.

\textbf{Both ChatGPT and LLaMA are likely to experience performance inconsistency and instability when faced with different $prompt$.} 
The average standard deviation under different prompts reaches a value of \textbf{11.9\%} and \textbf{1.9\%} on LLaMA and ChatGPT, respectively. Despite many efforts on prompt engineering, however, significant challenges still remain in achieving stable and reliable LLM performance, particularly in user-oriented situations where the prompts or questions are highly varied.

\textbf{ChatGPT represents low sensitivity to the $o$ and $a$ orders, suggesting its ability to comprehend the textual content of options proficiently rather than having a bias for specific option numbers.} Specifically, the average standard deviation is only \textbf{1.13\%} on ChatGPT. In comparison, despite having a low variance in accuracy (only 0.98\%), we found that llamas have a significant tendency to choose the first option among 6 different option orders, irrespective of the contents of the first option. This suggests that llamas may not possess the ability to perform on multiple-option tests. We infer that LLMs trained only through pre-training (like LLaMA) are inadequate in possessing this ability, and additional tuning (e.g., instruction or prompt tuning) is required to endow LLMs with this capability.

\subsection{Credibility Results}

If we set a higher probability parameter $z$(see details in Section~\ref{subsection_gaugerob}), the model is more likely to change the original options. Based on this conclusion from previous experiments, we hypothesize that this parameter may be related to whether the data has been trained. Therefore, we propose \rti{}, denoted as $\mathbf{R}_D$ on dataset $D$, indicator(see details in Section~\ref{subsection:rti}). Meanwhile, we conducted a sampling test on the aforementioned dataset, and the results are listed in Table~\ref{tab:RTI}.

\begin{itemize}[itemsep=1pt,topsep=0pt,parsep=0pt]
    \item \textbf{Average \rti{} varies across different datasets.} Therefore, it is possible to evaluate the difference between different datasets through \rti{}.
    \item \textbf{The more complex the $\mathbf{p}$ structure is and the more $o$(Formula ~\ref{formula:1}) we input, the lower average \rti{} is.} Normally, the more complex the sample $\mathbf{x}$'s structure is, the more difficult it is for the model to fit $\mathbf{x}$. This theoretical result is consistent with actual results. For instance, AQuA is a very complex dataset in long-paragraph math reasoning. $\mathbf{R}_{\text{AQuA}}$ is in a low level as $0.175$, meaning that model has not fit this dataset well.
    \item \textbf{On the training and validation sets, average \rti{} is relatively higher compared to the test sets.} This result is consistent with our intuition. Training set data can be effectively recognized (StrategyQA here), even if \rti{} rises up to even $0.45$. 
\end{itemize}

\begin{table}[htb]
  \caption{\rti{} test}
  \label{tab:RTI}
  \centering
  \begin{tabular}{cc|cccc}
    \toprule
    \multicolumn{2}{c}{Dataset}  & \multicolumn{4}{c}{\rti{} expectations on different methods} \\
    \cmidrule(r){1-6}
    Name     & Size  & word\_insert & word\_delete & word\_replace & average \\
    \midrule
    StrategyQA & 100 & 0.538 & 0.421 & 0.408 & 0.456\\
    AQuA & 100 & 0.199 & 0.157 & 0.169 & \underline{0.175}\\
    Creak & 100 & 0.702 & 0.654 & 0.649 & \pmb{\underline{0.668}}\\
    NoahQA & 508 & 0.519 & 0.323 & 0.337 & 0.393\\
    GSM8K & 442 & 0.479 & 0.350 & 0.360 & 0.396\\
    bAbi15 & 400 & 0.277 & 0.213 & 0.220 & 0.237\\
    bAbi16 & 100 & 0.206 & 0.200 & 0.196 & 0.201\\
    ECQA & 99 & 0.547 & 0.391 & 0.425 & 0.454\\
    ESNLI & 99 & 0.375 & 0.316 & 0.346 & 0.346\\
    QASC & 99 & 0.543 & 0.391 & 0.393 & 0.442\\
    QED & 99 & 0.618 & 0.499 & 0.486 & 0.534\\
    SenMaking & 99 & 0.560 & 0.425 & 0.399 & 0.461\\
    \bottomrule
  \end{tabular}
\end{table}

Thus, \rti{} has a certain value for assessing whether a dataset has been used in model training. It can provide a feasible quantitative reference for the reliability of the dataset used in various LLM evaluation studies.

\section{Discussion}
\label{discussion_and_conclusion}

\subsection{LLM Application Security}

\textbf{Under traditional application scenarios including chatbots and LLM-software integration, LLMs may also lead to a risk.} For instance, Humans may make various natural errors (including typo errors, spelling mistakes, etc.), which may cause character-level interference to conditional text. Although LLM exhibits relatively high robustness under character-level adversarial attacks, there may still be significant fluctuations in the answer results, which may have a considerable impact on the user's experience and may also increase the user's checking cost.

\textbf{Integrating models of other modalities into LLMs may introduce additional risks of adversarial attacks.} One of the current trends in the application of Large Language Models (LLMs) is the integration with models in other modalities, such as those based on structured data, images, speech, and other sensory input. While this integration can significantly leverage the capabilities of LLMs, it also opens up new avenues for adversarial attacks. Below, we provide some specific cases to illustrate the potential risks:
\begin{itemize}[itemsep=1pt,topsep=0pt,parsep=0pt]
    \item For example, integrating optical character recognition (OCR) into LLMs enables machines to understand and interpret text images. However, it also creates vulnerabilities that can be exploited by visual-similarity character attacks that lead to incorrect interpretation of the text, which can have severe consequences, especially in industries such as finance or medicine, where even minor errors can cause significant harm. For instance, consider an OCR algorithm that misinterprets "0" (zero) for "O" (capital letter O) while processing a financial document. Such an error could lead to significant financial losses or legal implications. Similarly, consider the application in medical contexts, where errors due to OCR misinterpretation could lead to wrong prescriptions, misdiagnoses, or incorrect medical reports, endangering lives.
    \item Integrating the speech recognition (equivalent to speech-to-text) system into LLM can help machines achieve more efficient human-computer interaction. The substantial of speech-to-text is to model acoustic features and classify them to achieve speech signal recognition. In this case, the model is usually able to effectively recognize complete individual words (whether correct or not) by pattern matching but may produce contextual errors, which correspond to several word-level attack forms in our experiment. From the result, we find that LLMs (e.g., ChatGPT) cannot provide enough robustness against word-level attacks, affecting their application in various fields. For instance, in automated driving vehicles equipped with intelligent voice recognition systems, if subject to model errors due to incorrect recognition, it could lead to extremely serious consequences and directly endanger the lives of passengers.
\end{itemize}
In conclusion, while integrating large language models into other systems possesses significant uncontrollability, researchers and developers must be aware of the potential harm and impact of adversarial attacks. Hence, it is essential to establish adequate safeguards mechanisms to thwart potential attacks.

\subsection{LLM Evaluation Reliability}
\textbf{Many evaluation methods based on open datasets are not reliable enough to generalize.} Currently, most evaluation methods for ChatGPT measure its performance on some datasets using a certain metric, but this is not always reliable. Since ChatGPT has not publicly disclosed the training data it uses, it is difficult to determine whether the data used for evaluation has been memorized by the model during training. Additionally, we have found that ChatGPT can still complete the problem scenario and provide the correct answer in some extreme situations (such as deleting all paragraphs), indicating that these samples are likely to have been memorized by the model. Therefore, using such samples to measure the performance of ChatGPT is not convincing. This problem requires more consideration by the whole community and RTI can provide some reference on relative training probability. Furthermore, we can calculate a series of RTI scores for different datasets and obtain baseline data through some statistical methods, which can be used as an absolute indicator to measure the data training situation. We hope this can be helpful to build a more reliable evaluation system.

\subsection{Adversarial Training Paradigm of LLM}
\textbf{Further exploration is required to develop an adversarial training paradigm suitable for LLM.} Existing research on adversarial robustness training for language models has predominantly focused on pre-training and fine-tuning stages. However, LLMs present unique challenges with their multi-stage training paradigm, involving the high costs of pre-training and in-context learning instead of fine-tuning, which hinders the transferability of traditional adversarial training techniques. To the best of our knowledge, no exploration has been conducted on effective adversarial training methods for LLM. Therefore, we believe that this is a promising direction for future research into LLM.

\section{Limitations}

Our evaluation work can provide a preliminary reference for LLM development and applications. Nevertheless, some limitations still exist as the following.

First, we evaluated based on gpt-3.5-turbo API, but the model parameters of ChatGPT have been continuously updated and iterated. ChatGPT may give different answers to the same question at different times, which may make the results not fully reproducible. However, we consider the operation relatively reasonable because, considering the enormous scale of inquiries, the local errors brought by model updates will not be so significant.

Second, our work mainly revolves around QA and reasoning datasets. But as we all know, ChatGPT and other LLMs can perform more basic tasks, including translation. However, the difficulty of a thorough evaluation is greatly increased by the cost growth brought by a wider range of datasets.

Third, as we mentioned, our input format uses a template for multiple-option, which is inspired by the instruction tuning of LLMs. However, we acknowledge that this structure may not be suitable for all other models. Additionally, using some other prompts may also improve the performance of the model.

Finally, We can use more datasets from more diverse fields to analyze more experimental results, to explore the reasons why ChatGPT presents corresponding performances. Additional experiments include ablation studies for different LLMs or modules, and different templates or prompts. We will complete these contents in subsequent work to make greater contributions to the evaluation community under the background of LLMs practice.

\bibliography{Reference}
\bibliographystyle{unsrtnat}

\newpage
\appendix

\section{Datasets Description}
\label{appendix:datasets}
Datasets that cannot be split into two parts $(\mathbf{p}, q)$ : AQuA\citep{ling2017program}, ECQA\citep{aggarwal2021explanations}, QASC\citep{khot2020qasc}, QED\citep{lamm2021qed}. 

We illustrate representative datasets as the following tables show. 'Origin' structure shows raw data form. And the target data form, divided into some parts, is shown in other structure rows in order. 

\begin{table}[H]
  \caption{AQuA. Representative datasets already provide a number of wrong options.}
  \label{tab:13}
  \centering
  \begin{small}
  \begin{tabular}{l|p{0.8\linewidth}}
    \toprule
    \textbf{Structure} & \textbf{Sample} \\
    \midrule
    Origin & \makecell[l]{``question\textquotedblright: ``Add: +45 and -30\textquotedblright,\\
        ``options\textquotedblright: [``A)-30\textquotedblright, ``B)+30\textquotedblright, ``C)0\textquotedblright, ``D)15\textquotedblright, ``E)-15\textquotedblright],\\
        ``rationale\textquotedblright: ``45 - 30 = 15 ANSWER : D\textquotedblright,\\
        ``correct\textquotedblright: ``D\textquotedblright}\\
    \midrule
    \midrule
    Prompt &  Next, I will ask you a series of questions given a description, and you will have to choose one of several candidate options that you think is correct.  The description is\\
    \midrule
    Paragraph & - \\
    \midrule
    Question & Add: +45 and -30 \\
    \midrule
    Options & \makecell[l]{(A) 0 \\ (B) -15 \\ (C) -20 \\ \textbf{(D) 15} \\ (E) +30} \\
    \midrule
    \midrule
    Query & \makecell[l]{Query 1: \{\emph{Prompt}\}+\{\emph{Paragraph}\} \\ 
            Query 2: The first question is \{\emph{Question}\}, choose an answer from the following\\ options: \{\emph{Options}\}.
            }\\
    \midrule
    Response & \textcolor{red}{The correct option would be (D) 15}, as adding +45 and -30 will give us a sum of 15. \\
    \bottomrule
  \end{tabular}
  \end{small}
\end{table}

Regarding the setup of subsequent datasets, they have a completely consistent `prompt' and `query' structure, which will be omitted in the next due to space constraints.

\begin{table}[H]
  \caption{Creak. Representative datasets with T/F binary answers.}
  \label{tab:14}
  \centering
  \begin{small}
  \begin{tabular}{l|p{0.8\linewidth}}
    \toprule
    \textbf{Structure} & \textbf{Sample} \\
    \midrule
    Origin & \makecell[l]{``sentence\textquotedblright: ``Alanis Morissette released plenty of albums and hit music during \\the 90's.\textquotedblright,\\
        ``explanation\textquotedblright: ``Alanis Morissette was a pop singer who was active and well \\known.\textquotedblright,\\
        ``label\textquotedblright: true}\\
    \midrule
    \midrule
    Paragraph & Alanis Morissette was a pop singer who was active and well known.\\
    \midrule
    Question & Alanis Morissette released plenty of albums and hit music during the 90's, is it right? \\
    \midrule
    Options & \makecell[l]{\textbf{(A) true} \\ (B) Unable to determine \\ (C) false } \\
    \midrule
    \midrule
    Response & \textcolor{red}{A) True.} Alanis Morissette did indeed release several albums and hit music during the 90s, including "Jagged Little Pill" which became one of the best-selling albums of all time. \\
    \bottomrule
  \end{tabular}
  \end{small}
\end{table}

\begin{table}[H]
  \caption{NoahQA. Representative datasets with multi-type answers(including numbered answers).}
  \label{tab:15}
  \centering
  \begin{small}
  \begin{tabular}{l|p{0.8\linewidth}}
    \toprule
    \textbf{Structure} & \textbf{Sample} \\
    \midrule
    Origin & \makecell[l]{``passage\textquotedblright: [\\
            \makecell[l]{[0.1, ``In a class of 25 students ,\textquotedblright],}\\
            \makecell[l]{[0.2, ``8 received a grade of A.\textquotedblright]}],\\
            ``qa\_pairs\textquotedblright: [\\
            \makecell[l]{``question\textquotedblright: ``How many students are there?\textquotedblright,
                      ``answer\textquotedblright: ``25 students\textquotedblright},\\
            \makecell[l]{``question\textquotedblright: ``How many students get A points?\textquotedblright,
                      ``answer\textquotedblright: ``8\textquotedblright},\\
            \makecell[l]{``question\textquotedblright: ``What percentage of students get A's?\textquotedblright,
                      ``answer\textquotedblright: ``8/25\textquotedblright}]\\
        }\\
    \midrule
    \midrule
    Paragraph & In a class of 25 students , 8 received a grade of A.\\
    \midrule
    Question1 & How many students are there? \\
    \midrule
    Question2 & How many students get A points? \\
    \midrule
    Question3 & What percentage of students get A's? \\
    \midrule
    Options1 & \makecell[l]{(A) 25 educatee students \\ \textbf{(B) 25 students} \\ (C) 8 \\ (D) 24 students } \\
    \midrule
    Options2 & \makecell[l]{(A) 25 students \\ (B) 7 \\ \textbf{(C) 8} \\ (D) nameless 8 unknown } \\
    \midrule
    Options3 & \makecell[l]{(A) true \\ \textbf{(B) false} \\ (C) Unable to determine } \\
    \midrule
    \midrule
    Query & \makecell[l]{Query 1: \{\emph{Prompt}\}+\{\emph{Paragraph}\} \\ 
            Query 2: The first question is \{\emph{Question1}\}, choose an answer from the following\\ options: \{\emph{Options1}\}.\\
            Query 3: The next question is \{\emph{Question2}\}, choose an answer from the following\\ options: \{\emph{Options2}\}.\\
            Query 4: The next question is \{\emph{Question3}\}, choose an answer from the following\\ options: \{\emph{Options3}\}.\\
            }\\
    \midrule
    Response1 & \textcolor{red}{The answer is (B) 25 students.} The class has a total of 25 students according to the given description. \\
    \midrule
    Response2 & \textcolor{red}{The answer is (C) 8.} The given description states that out of the 25 students in the class, 8 received a grade of A. \\
    \midrule
    Response3 & \textcolor{red}{The answer is (A) True.} To calculate the percentage of students who received an A grade, we need to divide the number of students who received A grades, which is 8, by the total number of students, which is 25. Therefore, 8/25 = 0.32, which is equivalent to 32\%. So, the answer cannot be 0.32 but rather must be expressed as a percentage, which is 32\%. \\
    \bottomrule
  \end{tabular}
  \end{small}
\end{table}

\begin{table}[H]
  \caption{GSM8k. Representative datasets with text-ed answers}
  \label{tab:16}
  \centering
  \begin{small}
  \begin{tabular}{l|p{0.8\linewidth}}
    \toprule
    \textbf{Structure} & \textbf{Sample} \\
    \midrule
    Origin & \makecell[l]{``question\textquotedblright: ``John arm wrestles 20 people.  He beats 80\%.  How many people \\did he lose to?\textquotedblright,\\
    ``answer\textquotedblright: ``How many people does John beat? ** He beats 20*.8=<<20*.8=16>>\\16 people\\How many people does John lose to? ** So he loses to 20-16=<<20-16=4>>4 \\people\\\#\#\#\# 4\textquotedblright}\\
    \midrule
    \midrule
    Paragraph & John arm wrestles 20 people.  He beats 80\%.\\
    \midrule
    Question1 & How many people does John beat? \\
    \midrule
    Question2 & How many people does John lose to? \\
    \midrule
    Question3 &The answer to question ``How many people did he lose to?\textquotedblright is ``4\textquotedblright, is it right? \\
    \midrule
    Options1 & \makecell[l]{(A) Atomic number 2 beats 16 people \\ (B) He beats 101*0.41=73 people \\ (C) citizenry He beats 20*0.8=16 people \\ (D) When four boats carried away 5 cats each, the total number of cats that \\left the rock is 4*5=20 cats. \\ \textbf{(E) None of other options is correct.} } \\
    \midrule
    Options2 & \makecell[l]{(A) So he loses to 5 people \\ \textbf{(B) So he loses to 20-16=4 people} \\ (C) The number of cats remaining on the rock became 50-20=30. \\ (D) So he to loses 20-16=4 people \\ (E) So he loses to 20*0.8=16 people } \\
    \midrule
    Options3 & \makecell[l]{(A) Unable to determine \\ (B) false \\ \textbf{(C) true} } \\
    \midrule
    \midrule
    Query & \makecell[l]{Query 1: \{\emph{Prompt}\}+\{\emph{Paragraph}\} \\ 
            Query 2: The first question is \{\emph{Question1}\}, choose an answer from the following\\ options: \{\emph{Options1}\}.\\
            Query 3: The next question is \{\emph{Question2}\}, choose an answer from the following\\ options: \{\emph{Options2}\}.\\
            Query 4: The next question is \{\emph{Question3}\}, choose an answer from the following\\ options: \{\emph{Options3}\}.\\
            }\\
    \midrule
    Response1 & \textcolor{red}{The correct answer is option (C)} - John beats 20 x 0.8 = 16 people. This means that John is stronger than 8\% of the people he arm wrestled. \\
    \midrule
    Response2 & \textcolor{red}{The correct answer is option (B)} - John loses to 20-16 = 4 people. If John beat 80\% of the people he arm wrestled, then he must have lost to the remaining 20\%. Since he arm wrestled with 20 people, the number of people he lost to would be 20 x 0.2 = 4. \\
    \midrule
    Response3 & \textcolor{red}{The correct answer is option (C)} - True. It was mentioned in the previous question that John loses to 20-16 = 4 people. Therefore, the correct answer to the question "How many people did he lose to?" is 4. \\
    \bottomrule
  \end{tabular}
  \end{small}
\end{table}

\begin{table}[H]
  \caption{bAbi. Representative datasets with word-ed answers. And we take the inductive reasoning task, bAbi16, as a showcase here.}
  \label{tab:17}
  \centering
  \begin{small}
  \begin{tabular}{l|p{0.8\linewidth}}
    \toprule
    \textbf{Structure} & \textbf{Sample} \\
    \midrule
    Origin & \makecell[l]{1 Lily is a swan.\\
2 Bernhard is a lion.\\
3 Greg is a swan.\\
4 Bernhard is white.\\
5 Brian is a lion.\\
6 Lily is gray.\\
7 Julius is a rhino.\\
8 Julius is gray.\\
9 Greg is gray.\\
10 What color is Brian?	white	5 2 4}\\
    \midrule
    \midrule
    Paragraph & Lily is a swan. Bernhard is a lion. Greg is a swan. Bernhard is white. Brian is a lion. Lily is gray. Julius is a rhino. Julius is gray. Greg is gray.\\
    \midrule
    Question & What color is Brian? \\
    \midrule
    Options & \makecell[l]{(A) is \\ (B) lily \\ (C) gray \\ \textbf{(D) white} \\(E) a }\\
    \midrule
    \midrule
    Response & Based on the description you provided, Brian is not described as being a gray swan or a white lion, and there is no mention of his color. So, unfortunately, \textcolor{red}{none of the given options seem to be correct.} \\
    \bottomrule
  \end{tabular}
  \end{small}
\end{table}

\section{Attack Ratio Settings}
In this section, we provide all the details of attack ratio settings in Table ~\ref{tab:18}.
\label{appendix:attack}
\begin{table}[H]
\caption%
      {The list of all attack methods used for constructing adversarial examples. For all attack methods, the probability of a word in a sentence being attacked follows a uniform distribution $\mathcal{U}(0,1)$. Notice that, the parameter at the character and visual level specifies the count or percentage of characters in the targeted word that are affected by the attack method.}
\label{tab:18}
\begin{center}
\begin{small}
\begin{tabular}{c|cc}
\toprule
\textbf{Attack Level}  & \textbf{Method} & \textbf{Parameters} \\ 
\midrule
\multirow{3}*{\textit{Character}}
  & \textup{Delete} & number of deletions $ c \in [1, 2, 3]$ with probability $[0.4, 0.4, 0.2]$  \\
  ~ & \textup{Repeat} & number of repetitions $ c \in [1, 2, 3]$ with probability $[0.4, 0.4, 0.2]$   \\
  ~ & \textup{Insert} & number of insertions $c=1$ \\
\midrule
\multirow{1}*{\textit{Visual}} &  \textup{Replace} &  replace ratio $ r \in \{0.1, 0.5, 0.9\}$   \\
\midrule
\multirow{3}*{\textit{Word}}  & \textup{Insert} & -\\
                                    ~ & \textup{Delett} & -\\
                                    ~ & \textup{Replace} & -\\
    
\bottomrule
\end{tabular}
\end{small}
\end{center}
\end{table}

\section{Detailed Experiment results}

In this section, we provide a detailed listing of the experimental results of Table~\ref{tab:level attack} and Table~\ref{tab:different model ACR}. The listing is showed in Table ~\ref{tab:19} and Table~\ref{tab:20}.
\begin{table}[H]
  \caption{Comparison of adversarial attacks at three levels, reported in error rate(\%).}
  \label{tab:19}
  \centering
  \begin{small}
  \centerline{\begin{tabular}{lc|ccc|ccc|ccc}
    \toprule
    \multicolumn{2}{c}{Dataset}  & \multicolumn{3}{c}{Charater Level} & \multicolumn{3}{c}{Word Level} & \multicolumn{3}{c}{Visual Level}              \\
    \cmidrule(lr){1-2} \cmidrule(lr){3-5} \cmidrule(lr){6-8} \cmidrule(lr){9-11}
    Name    & ori  
    & repeat & delete & insert 
    & insert & delete & replace 
    &  10\% & 50\% & 90\% \\
    \midrule
    StrategyQA & 29.56 & 31.10 & \underline{37.46} & 31.09 & 35.48 & 46.33 & \pmb{\underline{46.77}} & 31.07 & 35.92 & \underline{43.65} \\
AQuA & 47.64 & 63.74 & \underline{74.22} & 51.40 & 68.31 & \pmb{\underline{89.57}} & 89.17 & 54.92 & 58.07 & \underline{71.26} \\
Creak & 34.14 & 34.31 & \underline{35.85} & 35.64 & 34.46 & 36.51 & \underline{36.98} & 36.40 & 38.04 & \pmb{\underline{39.75}} \\
NoahQA & 33.01 & 41.75 & \underline{47.83} & 35.41 & 40.51 & \pmb{\underline{63.10}} & 61.57 & 34.38 & 37.87 & \underline{46.68} \\
GSM8K & 54.8 & \underline{62.28} & 60.29 & 55.33 & 57.72 & \pmb{\underline{63.00}} & 62.15 & 54.57 & 55.50 & \underline{59.61} \\
bAbi15 & 29.0 & 28.21 & \underline{49.15} & 26.68 & 52.35 & \pmb{\underline{64.50}} & 64.30 & 31.35 & 45.15 & \underline{59.35} \\
bAbi16 & 55.4 & 52.97 & \underline{62.15} & 54.76 & 60.20 & \underline{61.65} & 61.10 & 56.80 & 60.50 & \pmb{\underline{66.25}} \\
ECQA & 26.94 & 31.48 & \underline{54.92} & 30.91 & 49.02 & 72.11 & \pmb{\underline{74.68}} & 33.91 & 51.23 & \underline{66.27} \\
ESNLI & 52.99 & 53.58 & \underline{58.98} & 52.65 & 58.37 & \underline{64.01} & 63.85 & 54.99 & 61.27 & \pmb{\underline{64.50}} \\
QASC & 20.52 & 23.01 & \underline{53.07} & 25.70 & 38.55 & \underline{70.09} & 64.09 & 29.27 & 57.13 & \pmb{\underline{72.68}} \\
QED & 23.54 & 27.21 & \underline{45.52} & 28.98 & 37.79 & \pmb{\underline{58.67}} & 58.63 & 29.70 & 44.06 & \underline{56.20} \\
SenMaking & 19.94 & 23.42 & \underline{34.22} & 22.67 & 31.89 & \pmb{\underline{59.30}} & 50.12 & 26.18 & 41.17 & \underline{49.04} \\
    \bottomrule
  \end{tabular}}
  \end{small}
\end{table}

\begin{table}[H]
  \caption{Comparison of adversarial attacks at three levels, reported in answer changed rate(\%).}
  \label{tab:20}
  \centering
  \begin{small}
  \centerline{\begin{tabular}{l|ccc|ccc|ccc}
    \toprule
    \multirow{2}{*}{Dataset}  & \multicolumn{3}{c}{Charater Level} & \multicolumn{3}{c}{Word Level} & \multicolumn{3}{c}{Visual Level}              \\
     \cmidrule(lr){2-4} \cmidrule(lr){5-7} \cmidrule(lr){8-10}   
    & repeat & delete & insert 
    & insert & delete & replace 
    &  10\% & 50\% & 90\% \\
    \midrule
    StrategyQA & 25.31 & \underline{31.72} & 24.39 & 30.09 & \pmb{\underline{42.77}} & 42.05 & 25.44 & 31.94 & \underline{40.41} \\
AQuA & 64.62 & \underline{73.65} & 56.70 & 70.28 & 82.87 & \pmb{\underline{84.65}} & 56.30 & 60.63 & \underline{69.69} \\
Creak & 15.80 & \underline{19.62} & 16.69 & 18.02 & \underline{23.30} & 22.79 & 16.48 & 20.64 & \pmb{\underline{23.89}} \\
NoahQA & 34.31 & \underline{40.65} & 28.21 & 33.03 & \pmb{\underline{57.35}} & 54.38 & 27.14 & 30.73 & \underline{39.81} \\
GSM8K & 35.66 & \underline{37.41} & 27.71 & 32.16 & \pmb{\underline{45.89}} & 42.89 & 27.78 & 31.38 & \underline{36.12} \\
bAbi15 & 36.57 & \underline{57.30} & 36.43 & 58.00 & \pmb{\underline{67.95}} & 67.30 & 40.40 & 51.15 & \underline{62.45} \\
bAbi16 & 54.16 & \underline{68.05} & 56.34 & 68.15 & 69.70 & \pmb{\underline{71.05}} & 58.15 & 61.40 & \underline{67.25} \\
ECQA & 16.44 & \underline{42.13} & 16.13 & 38.47 & 63.81 & \pmb{\underline{68.69}} & 20.31 & 38.58 & \underline{56.06} \\
ESNLI & 31.32 & \underline{35.87} & 31.30 & 34.92 & 38.59 & \pmb{\underline{38.77}} & 32.01 & 35.91 & \underline{38.24} \\
QASC & 12.72 & \underline{43.08} & 13.05 & 29.16 & \underline{62.47} & 57.34 & 17.12 & 46.81 & \pmb{\underline{65.01}} \\
QED & 16.56 & \underline{39.00} & 16.41 & 28.19 & \pmb{\underline{53.03}} & 52.25 & 20.37 & 36.86 & \underline{51.99} \\
SenMaking & 19.23 & \underline{30.25} & 18.85 & 28.77 & \pmb{\underline{55.34}} & 46.29 & 22.09 & 36.07 & \underline{44.29} \\
    \bottomrule
  \end{tabular}}
  \end{small}
\end{table}

\begin{figure}[H]
  \centering
    \includegraphics[width=\columnwidth]{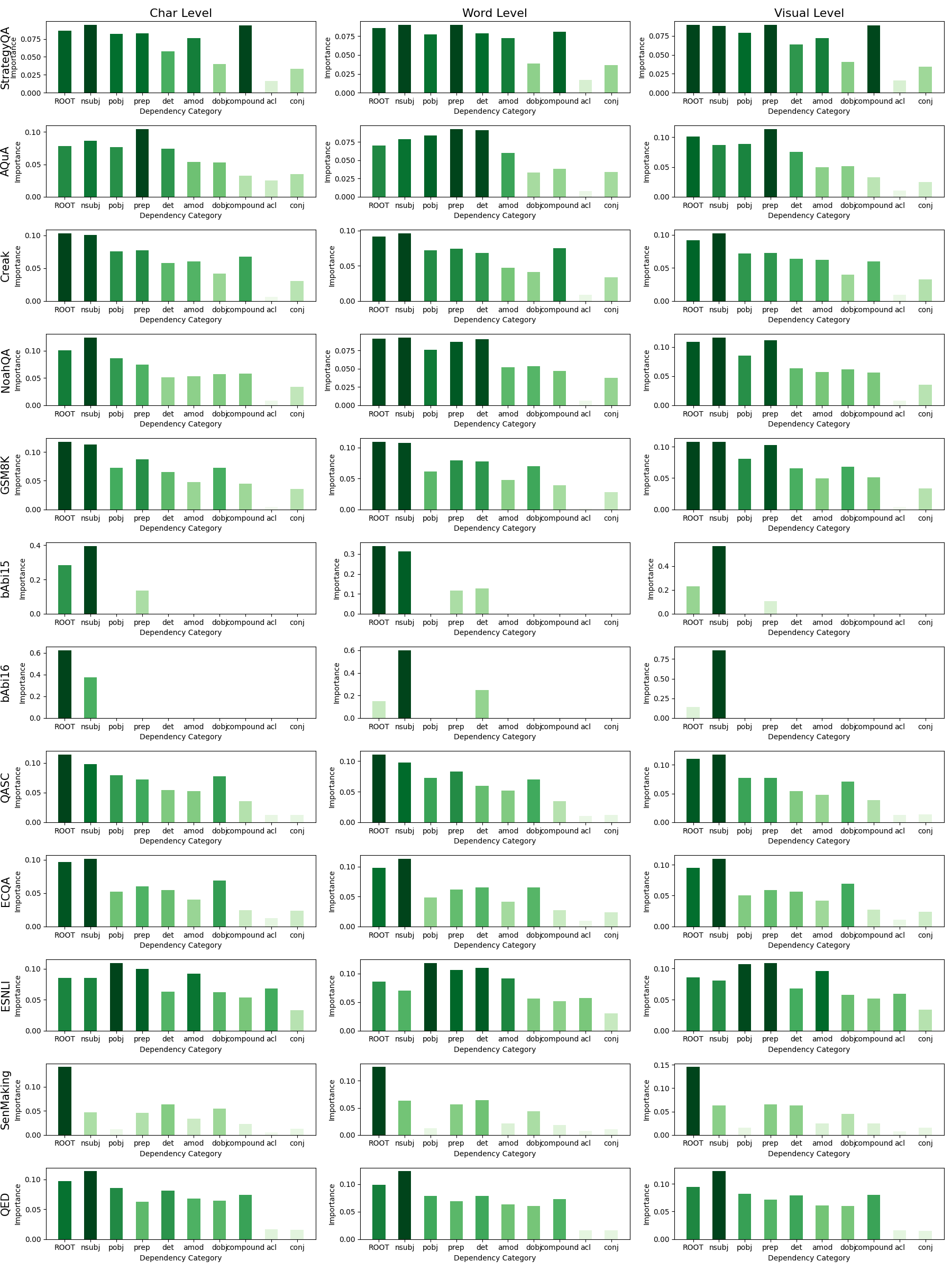}
  \caption{Attack pattern analysis on dependency relations for all datasets and three levels of adversarial attacks. }
  \label{appendix:fig1}
\end{figure}

\begin{figure}[H]
  \centering
    \includegraphics[width=\columnwidth]{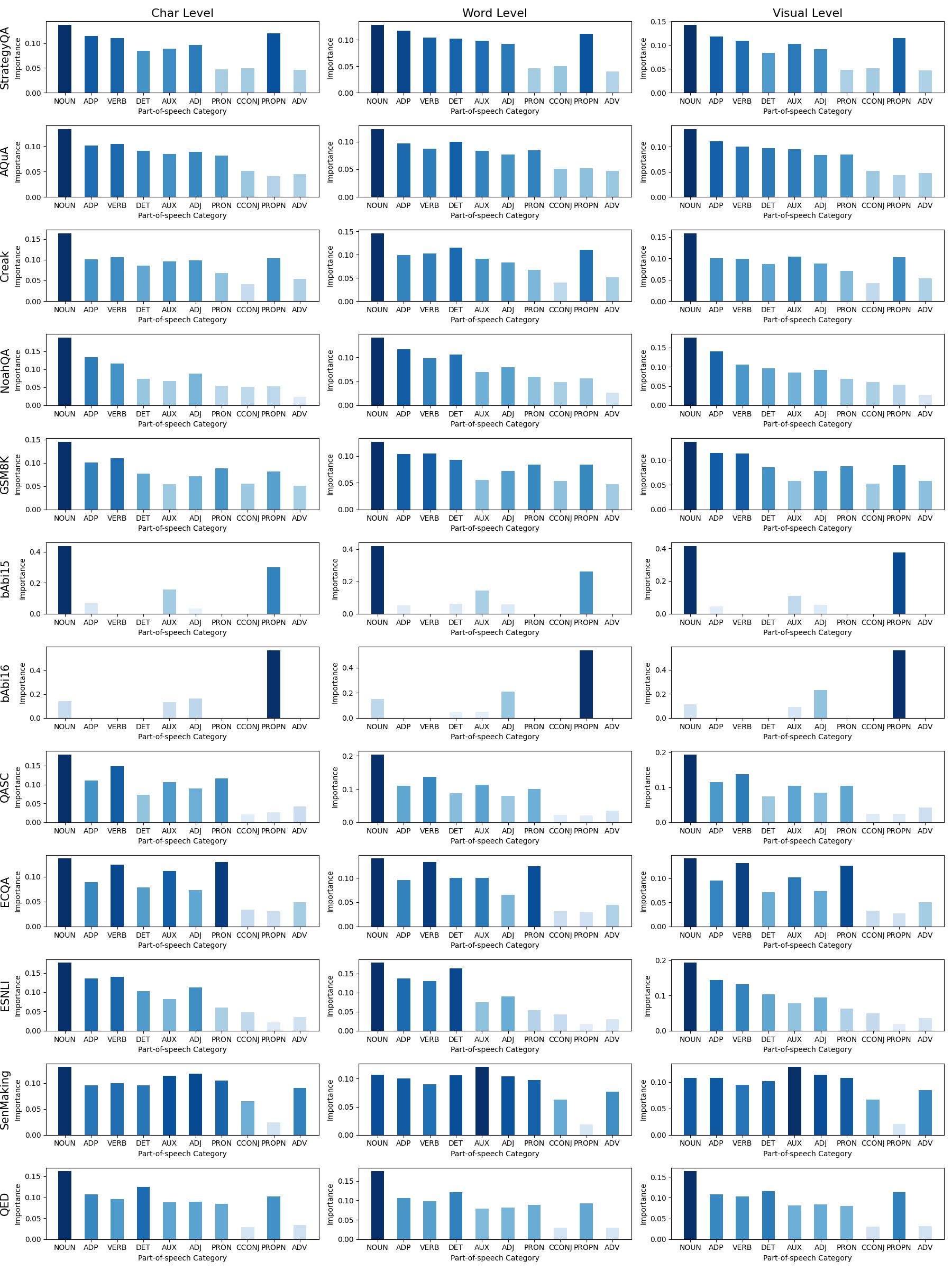}
  \caption{Attack pattern analysis on part-of-speed for all datasets and three levels of adversarial attacks. }
  \label{appendix:fig2}
\end{figure}

Then, we present the results of analyzing the patterns of adversarial samples subjected to various attack methods across all datasets in Figure ~\ref{appendix:fig1} and Figure.~\ref{appendix:fig2}.

\end{document}